\documentclass{article} % For LaTeX2e
\usepackage{iclr2027_conference,times}

\usepackage{amsmath,amsfonts,bm}

\def\eqref#1{equation~\ref{#1}}
\def\1{\bm{1}}

\DeclareMathAlphabet{\mathsfit}{\encodingdefault}{\sfdefault}{m}{sl}
\SetMathAlphabet{\mathsfit}{bold}{\encodingdefault}{\sfdefault}{bx}{n}

\usepackage{hyperref}
\usepackage{url}
\usepackage{booktabs}  
\usepackage{microtype}      % microtypography
\usepackage{xcolor}         % colors
\usepackage{graphicx}
\usepackage{wrapfig}
\usepackage{colortbl}
\usepackage{xcolor}

\usepackage{subcaption}
\usepackage{multirow}
\usepackage{multicol}
\usepackage[most]{tcolorbox}
\usepackage{cleveref}
\definecolor{darkgreen}{rgb}{0.0, 0.5, 0.0}
\usepackage{cleveref}
\crefname{figure}{Fig.}{Figs.}
\Crefname{figure}{Fig.}{Figs.}

\crefname{section}{Sec.}{Secs.}
\Crefname{section}{Sec.}{Secs.}

\newcommand{\methodName}{\textsc{SpatialUncertain}}

\title{Seeing Isn't Knowing: Do VLMs Know When Not to Answer Spatial Questions (and Why)? }

\author{
\hspace{-1em}
\textbf{
Yue Zhang$^{1}$ \hspace{-0.01em}
Zun Wang$^{1}$ \hspace{-0.01em}
Han Lin$^{1}$ \hspace{-0.01em}
Yonatan Bitton$^{2}$ \hspace{-0.01em}
Idan Szpektor$^{2}$ \hspace{-0.01em}
Mohit Bansal$^{1}$
} \\
$^{1}$UNC Chapel Hill
\qquad
$^{2}$Google Research \\
\href{https://zhangyuejoslin.github.io/spatialuncertain/}{\texttt{https://spatialuncertain.github.io}}
}

\iclrfinalcopy % Uncomment for camera-ready version, but NOT for submission.
\begin{document}

\maketitle

\begin{abstract}

Spatial reasoning benchmarks typically evaluate whether vision-language models can derive the correct answer from a visual observation. Yet in real 3D environments, the observation itself may be unreliable: occlusion can remove task-relevant evidence, while perspective can make visible geometry misleading. Reliable spatial reasoning therefore requires more than answering a question correctly. A model must also assess whether its current observation provides sufficient and trustworthy evidence for that answer.
We introduce \methodName, a controlled evaluation framework for studying \emph{viewpoint-dependent observational uncertainty}. By holding the underlying 3D scene fixed while systematically varying the observation, we study two complementary failure modes: missing evidence caused by occlusion and misleading evidence caused by perspective. We further evaluate whether models can recognize when the current view is unreliable and identify a more informative observation.
Across eight open- and closed-source vision-language models from the Qwen2.5-VL, InternVL3, GPT, and Gemini families, we find that model behavior does not track the reliability of visual evidence.
Models do not reliably become more cautious as evidence disappears, and under perspective conflict, their judgments increasingly follow projected appearance rather than the unchanged physical 3D relation. 
Internal analysis suggests a corresponding representational asymmetry: projected 2D relations are readily available, whereas the underlying physical 3D relation is barely decodable. Moreover, models that can identify an informative viewpoint when explicitly asked often fail to recognize when such an additional view is needed.
These failures are not fully resolved by prompting or fine-tuning, and providing a better viewpoint is substantially more effective than adding depth information to the same misleading observation. Our results identify assessing the reliability of visual observations as a distinct and missing component of current spatial reasoning evaluation.

\end{abstract}
\section{Introduction}

\begin{figure}
    \centering
\includegraphics[width=\linewidth]{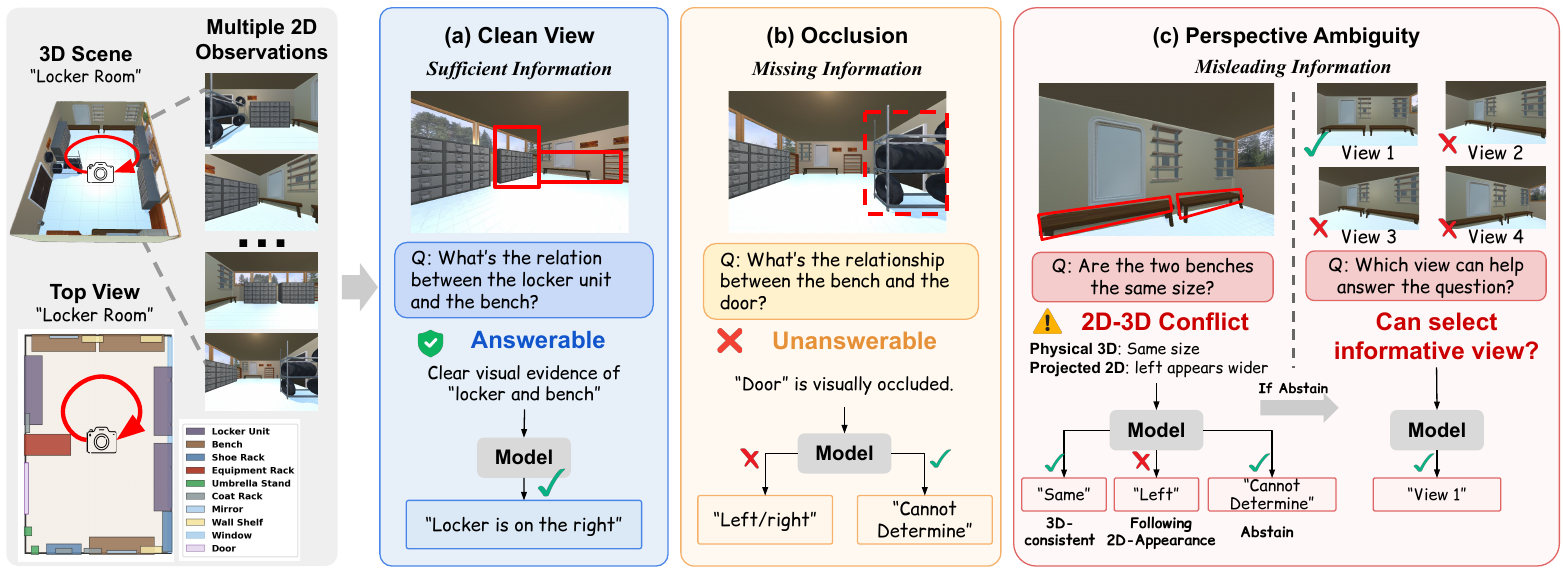}
  \caption{
Visual observations are inherently 2D projections of a 3D world and can provide sufficient, missing, or misleading evidence for spatial reasoning. 
(a) Under clean views, spatial questions are answerable from direct visual evidence. 
(b) Under occlusion, task-relevant evidence may become invisible and require abstention with \textit{Cannot determine}. 
(c) Under perspective ambiguity, projected 2D appearance can conflict with the underlying physical 3D relation, requiring models to recognize conflicting evidence and identify a more informative view for reliable reasoning.
}
    \label{fig:teaser_fig}
    
\end{figure}

Recent advances in Multimodal Large Language Models (MLLMs)~\citep{liu2023visual, singh2025openai,gemini2.5flash2025} have enabled intelligent agents to perceive and interact with their environments, bringing us closer to practical embodied systems~\citep{zhang2024vision}. A fundamental capability underlying these systems is spatial reasoning, which has been extensively studied through a growing number of benchmarks~\citep{yang2025thinking, yang2025cambrian, pothiraj2025capture, wang2024picture, liu2025can, jia2025omnispatial, stogiannidis2025mind,wang2025spatial}. These benchmarks measure models’ ability to answer spatial questions, such as object relations, distance, and object size, from visual observations including images or videos. However, they typically assume that the provided visual observation contains sufficient and reliable evidence for answering the question~(see \Cref{fig:teaser_fig}(a)).

Visual observations are inherently 2D projections of a 3D world, and the spatial evidence they provide can become incomplete or misleading in different ways.
In the occlusion setting~(see~\Cref{fig:teaser_fig}(b)), a target object may become partially or fully hidden, making its spatial relations increasingly difficult or even impossible to determine from the current view. In the perspective setting~(see~\Cref{fig:teaser_fig}(c)), objects can remain fully visible while viewpoint changes distort their apparent geometry, potentially creating conflicts between projected 2D appearance and the underlying physical 3D relation.
Such unreliable observations are particularly challenging for embodied agents~\citep{zhang2024vision, duan2022survey, Zhang2023VLNTransTF, Yu2026WhenAH}, where acting on missing or misleading visual evidence can lead to incorrect action decisions or unsafe behaviors.
Ideally, a model should recognize when the current evidence is insufficient or unreliable, abstain when necessary, and seek additional observations when the current view does not support a reliable judgment.
A growing body of work studies uncertainty and abstention across language and multimodal models, focusing on whether a model can recognize when it does not know the answer~\citep{manakul2023selfcheckgpt,stengel2024lacie,wen2025know,madhusudhan2026knowing}.
A complementary challenge in visual spatial reasoning is whether the model can assess the reliability of the observation itself, namely, whether the available visual evidence is sufficient and trustworthy for making a spatial judgment.

To address this gap, we introduce \methodName, a controlled evaluation framework for studying \emph{viewpoint-dependent observational uncertainty} in spatial reasoning, where the reliability of visual evidence changes with the observation viewpoint while the underlying 3D scene remains fixed.
We construct paired observations with systematically varying degrees of evidence reliability, rather than a single binary distinction between answerable and unanswerable cases.
For occlusion, we progressively vary and quantify target visibility, producing clean, partially occluded, and fully occluded conditions that control how much task-relevant evidence remains observable. For perspective, we vary the camera viewpoint to induce different degrees of disagreement between projected 2D appearance and the corresponding physical 3D relation. We evaluate whether model behavior is appropriate to the evidence available in the current view. Under occlusion, this means answering when sufficient evidence remains and abstaining when it does not; under perspective conflict, we characterize whether predictions remain consistent with the physical 3D relation, follow the projected 2D appearance, or abstain.
For perspective cases, we further evaluate whether models can identify a more informative viewpoint through two complementary tasks: ViewSel, which measures viewpoint selection in isolation, and AbstainViewSel, which jointly tests whether a model can first recognize an unreliable observation and then select an alternative view that better supports the spatial judgment.

Using this controlled setup, we evaluate eight vision-language models spanning open-source (Qwen2.5-VL-7B, Qwen2.5-VL-32B, InternVL3-8B) and closed-source (GPT-4o, GPT-5-mini, GPT-5.4, Gemini-2.5-Flash, Gemini-3.0-Flash) families. Our results show that model behavior often fails to track the reliability of the available visual evidence (\Cref{sec:main result}). Under occlusion, abstention increases only weakly and inconsistently as target visibility decreases. Under perspective conflict, models increasingly follow the projected 2D appearance as disagreement with the underlying physical 3D relation becomes stronger. Although stronger models can often identify an informative viewpoint when explicitly asked, their performance drops substantially when they must first determine whether the current observation is reliable enough to answer or whether another view is needed.
We further diagnose these failures (\Cref{sec:why}). The physical 3D relation is barely decodable from model representations, while the projected 2D relation is readily available. Model attention is also biased toward the object that appears larger in projection. The same behavior persists under open-ended evaluation, and a real-image consistency check shows the same pattern: viewpoint changes can induce contradictory physical-size judgments despite unchanged object sizes.
Finally, we examine what helps models handle unreliable observations (\Cref{sec:improve}). Structured prompting improves performance on unanswerable cases but introduces a trade-off with answerable accuracy. Fine-tuning on diverse ambiguity conditions substantially reduces 2D-following, though largely by shifting predictions toward abstention rather than recovering the physical 3D relation. At inference time, providing an alternative viewpoint is more effective than augmenting the same misleading observation with depth information. Together, these results suggest that reliable spatial reasoning requires not only reasoning from visual observations, but also assessing when those observations should be trusted and when better evidence should be sought.

\vspace{-3mm}

\section{Related Work}

\noindent\textbf{Spatial reasoning in MLLMs.}
Spatial reasoning has emerged as a fundamental capability for multimodal large language models (MLLMs)~\citep{chen2024spatialvlm, Cheng2024SpatialRGPTGS, zhang2024spartun3d}, and a growing body of work has proposed benchmarks to evaluate it~\citep{yang2026cambrian,yang2025thinking, yu2026and, daxberger2025mm, wang2024embodiedscan, xu2025spatialbench, Rajabi2024GSRBENCHAB, yang2025mmsi, ma2022sqa3d}. Early efforts focus on evaluating basic spatial relations such as relative relations, depth ordering, and size comparison using image/video-based question answering datasets. More recent benchmarks aim to provide broader and more systematic evaluations, including multi-view~\citep{zhang2026multiview, wang2025spatial, zhang2026spinbench} and multi-task settings~\citep{yin2023large, jia2025omnispatial}, which cover diverse spatial reasoning skills.
Several works further emphasize the importance of controlled evaluation~\citep{pothiraj2025capture, liu2023visual, johnson2017clevr}. For example, What’sUp~\citep{kamath2023s} constructs minimally varying image pairs to isolate spatial relations. 
Despite these advances, existing benchmarks primarily evaluate whether models produce correct answers, but do not explicitly assess whether a question is answerable given the observation. 
In contrast, we evaluate spatial reasoning under varying observation conditions (e.g., occlusion and perspective ambiguity), focusing on whether models can recognize when the available evidence is reliable for spatial questions.

\noindent\textbf{Observational uncertainty and abstention.}
Uncertainty estimation and abstention are important for building reliable models. Classical work on calibration and selective prediction shows that neural networks can be overconfident, and that models should sometimes abstain when their predictions are uncertain~\citep{guo2017calibration,hendrycks2016baseline,geifman2017selective,whitehead2022reliable}. This problem has become especially important for large language models, which often generate fluent but unsupported answers~\citep{kirichenko2026abstentionbench}. Recent work therefore studies truthfulness, self-knowledge, confidence elicitation, hallucination detection, and calibrated expressions of uncertainty~\citep{lin2022truthfulqa,kadavath2022language,yin2023large,tian2023just,xiong2024can,manakul2023selfcheckgpt,stengel2024lacie,wen2025know}. 
While uncertainty and abstention have been widely explored in language models, they remain less studied in vision-language models. Related efforts examine object hallucination, unanswerable visual questions, and selective VQA, encouraging models to abstain rather than answer incorrectly~\citep{rohrbach2018object,li2023evaluating,sun2024aligning,guan2024hallusionbench,gurari2018vizwiz,guo2024unk,he2024tubench,eisenschlos2024selectively}. However, these works primarily focus on factual uncertainty, object existence, or generic unanswerability, and typically assume that visual observations provide reliable evidence for reasoning. 
In contrast, we study observation-dependent uncertainty in spatial reasoning, where answerability is determined by the viewpoint.

\section{\methodName: Controlled Evaluation Framework }

\begin{figure}
    \centering
    \includegraphics[width=\linewidth]{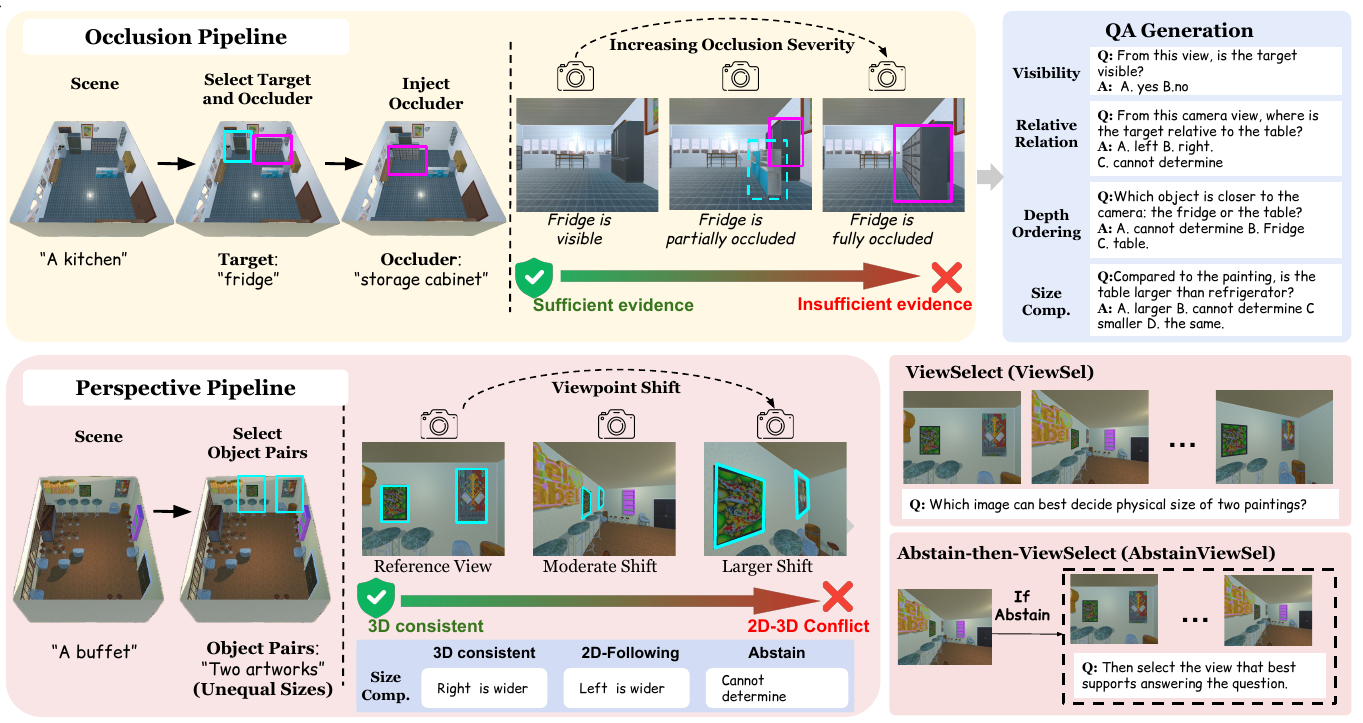}
    \caption{
Overview of our evaluation framework, \methodName.
\textbf{(Top) Occlusion:} We select a target-occluder pair and progressively decrease target visibility from a clean reference view to full occlusion. We evaluate four spatial question types: visibility, relative position, depth ordering, and size comparison.
\textbf{(Bottom) Perspective:} We select same-category object pairs and shift the camera viewpoint. We focus on horizontal or vertical size comparison and categorize model predictions as 3D-consistent, 2D-following, abstaining, or other.
We further introduce \textbf{ViewSel}, which directly evaluates informative-view selection, and \textbf{AbstainViewSel}, jointly evaluates whether a model can either reason correctly from the current view or, after abstaining, select a more informative one.
}
\vspace{-3mm}
    \label{fig:framework}
\end{figure}

We construct a controlled evaluation framework for studying viewpoint-dependent observational uncertainty in spatial reasoning, and the pipeline is shown in~\Cref{fig:framework}. 
Our key setting is that the underlying 3D indoor scene~(\Cref{3d scene collection}) remains fixed, while the reliability of the available visual evidence changes with the observation viewpoint. We instantiate two complementary forms of uncertainty: occlusion (\Cref{occlusion configuration}), where task-relevant evidence becomes missing from the current view, and perspective ambiguity (\Cref{perspective ambiguity}), where visual evidence remains available but its 2D projection may provide misleading cues about
the underlying 3D geometry.
All configurations undergo human validation to ensure quality (\Cref{quality control}).

\subsection{3D Scene Collection}
\label{3d scene collection}
We generate 3D indoor scenes using Holodeck~\citep{yang2024holodeck}, an LLM-based automated layout generation system. Given a natural language prompt (e.g., \textit{``a bedroom"} or \textit{``a kitchen"}), Holodeck uses a large language model (GPT-4o~\citep{gpt4o}) to plan object selection and placement, producing diverse and realistic room configurations. For each scene, Holodeck provides full 3D asset placement with object positions, orientations, and bounding box information, which we use to automatically derive ground truth annotations without the need for manual labeling.
All scenes are rendered using AI2-THOR~\citep{kolve2017ai2}, which supports controllable camera placement and produces photo-realistic RGB images. This controlled rendering environment is central to our framework: by fixing the 3D scene and varying only the camera viewpoint or object configuration, we can isolate how changes in observation reliability affect model reasoning.

\subsection{Occlusion Configurations}
\label{occlusion configuration}
\noindent\textbf{Target-occluder selection.}
For each clean scene, we select target objects satisfying two criteria: (1) \textit{visibility}: the object must be clearly visible from the camera viewpoint, measured by its projected bounding box area; and (2) \textit{uniqueness}: the object must be the only instance of its category in the scene, ensuring unambiguous reference in generated questions. We retain the top-k~(k=3) targets per scene ranked by visibility score (more details are in the Appendix~\ref{benchmark construction appendix}). 
In our setting, this procedure results in a diverse collection of target–occluder pairs, with targets spanning 225 unique object categories (e.g., bookshelf, coffee table, ottoman) and occluders spanning 286 categories (e.g., storage cabinet, bookshelf, armchair), covering a wide range of object types and occlusion scenarios.

\begin{figure}[t]
\centering

\begin{subfigure}[t]{0.49\linewidth}
    \centering
    \includegraphics[width=\linewidth]{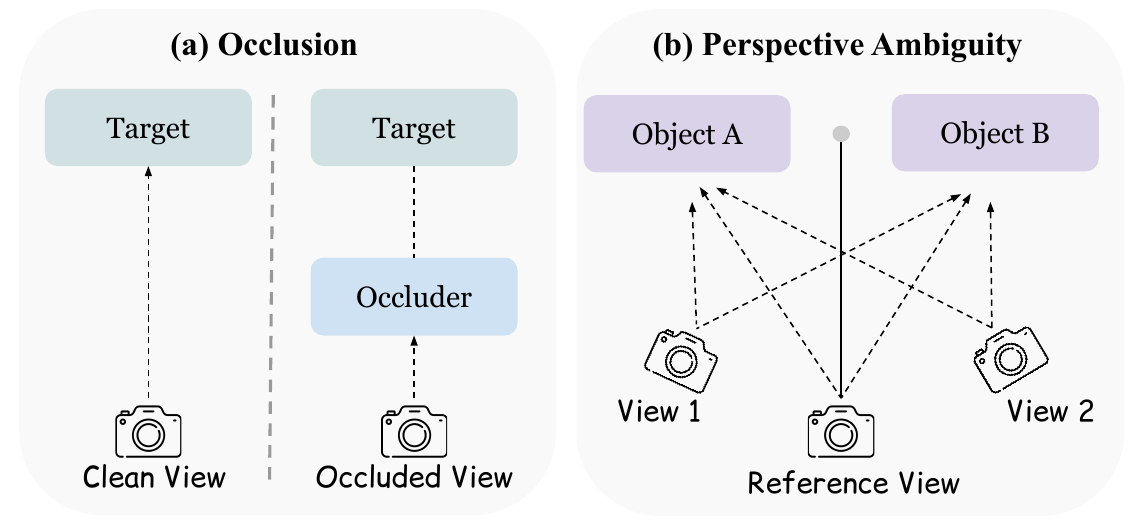}
    \caption{Camera placement under different conditions.}
    \label{fig:camera placement}
\end{subfigure}
\hfill
\begin{subfigure}[t]{0.48\linewidth}
    \centering
    \includegraphics[width=\linewidth]{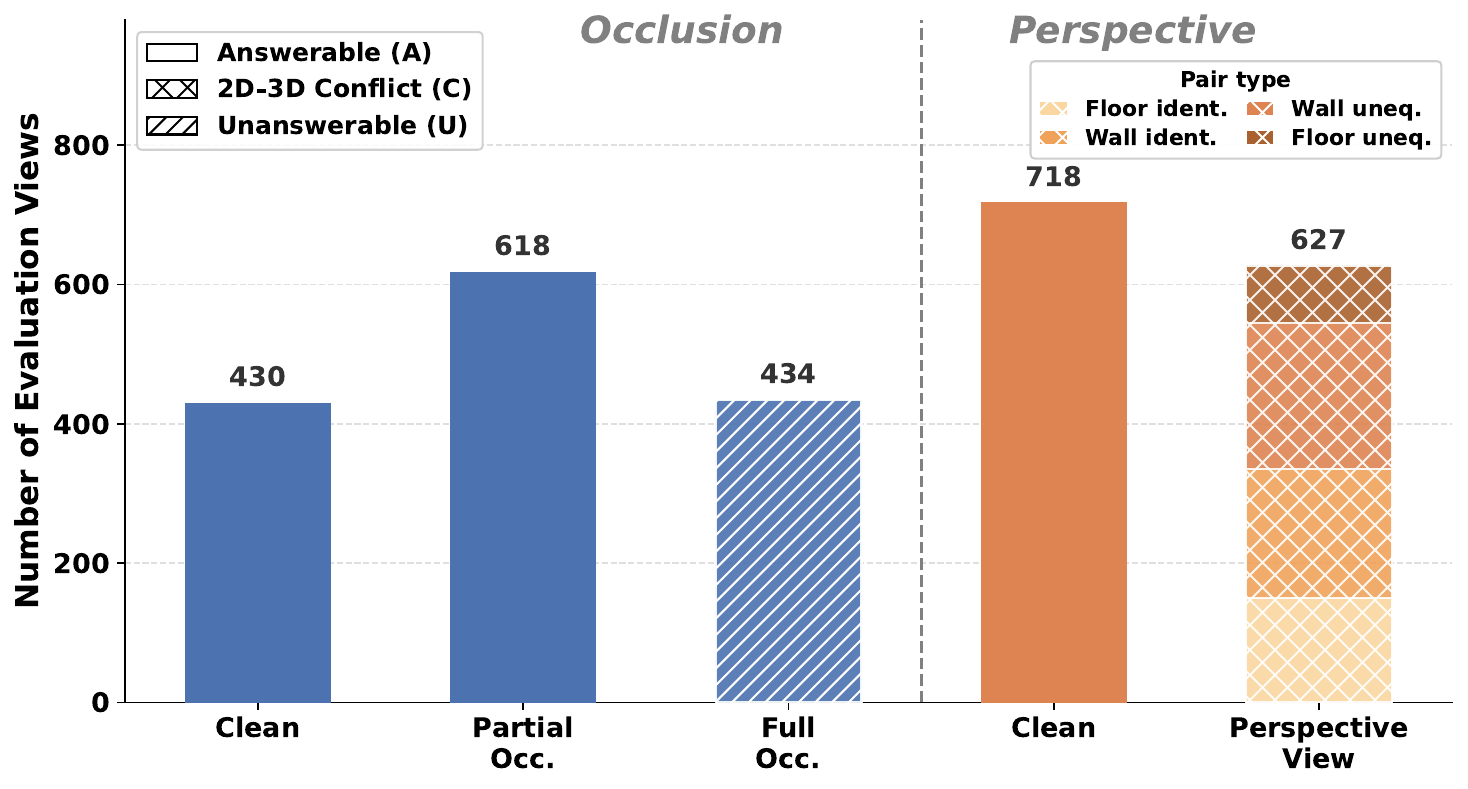}
    \caption{Distribution of types.}
    \label{fig:stats}
\end{subfigure}

\caption{Camera placement under different observation conditions (a) and question distributions for occlusion and perspective ambiguity (b), showing answerable/unanswerable cases under occlusion and 2D-3D conflict cases under perspective ambiguity.}
\vspace{-4mm}
\label{fig:answerability_overview}

\end{figure}

\noindent\textbf{Occlusion scene camera placement.}
Given a selected target-occluder pair, we place the occluder along the line of sight between the camera and the target, ensuring it is closer to the camera than the target (shown in ~\Cref{fig:answerability_overview}(a) left). 
The placement is subject to several geometric constraints to ensure physical plausibility: (1) the occluder must not penetrate other objects or room boundaries, (2) a minimum depth separation is maintained between the occluder and target to avoid clipping, and (3) the occluder must remain within the same room area as the target.

\noindent\textbf{Quantifying target visibility.}
For each target-occluder pair, we measure the fraction of the target that remains visible under occlusion using instance masks rendered from the same camera pose.
Let $M_{\mathrm{occ}}(T)$ denote the target mask in the original occluded scene, and $M_{\mathrm{ref}}(T)$ the target mask from a reference rendering in which only the designated occluder is hidden. 
We define the target visibility ratio as
$
r_{\mathrm{vis}}
=
\frac{|M_{\mathrm{occ}}(T)|}
{|M_{\mathrm{ref}}(T)|},
$
where $|\cdot|$ denotes the number of target pixels in the instance mask.
A smaller visibility ratio indicates stronger occlusion caused by the designated occluder, and we use this measure to group configurations by occlusion severity.
We categorize an occluded view as \emph{partial} when the measured target visibility satisfies \(r_{\mathrm{vis}} \geq 0.05\), and as \emph{full} when \(r_{\mathrm{vis}} < 0.05\).

\subsection{Perspective Ambiguity Configuration}
\label{perspective ambiguity}

\begin{figure}[t]
    \centering
    \includegraphics[width=\linewidth]{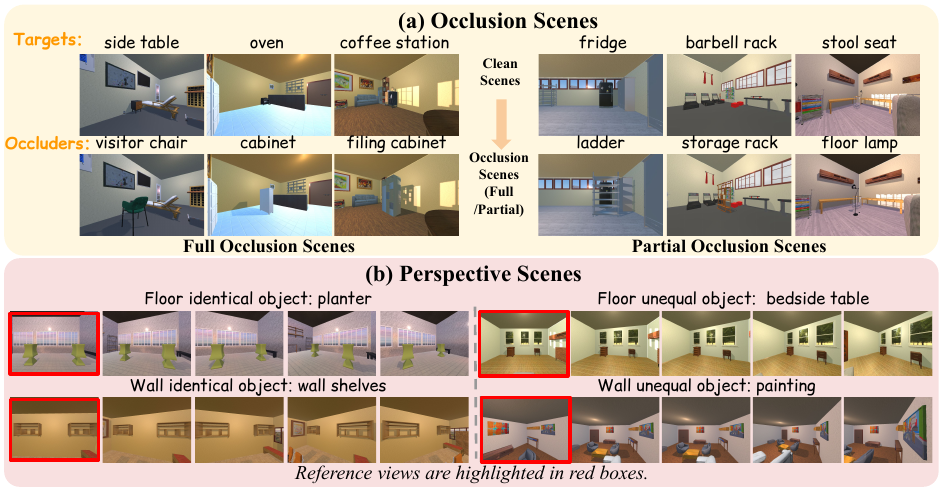}
\caption{
Examples of perspective scenes where camera shifts create conflicts between projected 2D appearance and the underlying 3D relation.
}
    \label{fig: perspective_scene}
    \vspace{-5mm}
\end{figure}

\noindent\textbf{Object pair selection.}
To study how viewpoint changes affect perceived object geometry, we select pairs of same-category objects whose physical dimensions can be compared unambiguously.
We consider two spatial configurations.
\emph{Floor pairs} consist of two floor-standing objects of the same category (e.g., two chairs) that are spatially proximate and have similar orientations.
\emph{Wall pairs} consist of two wall-mounted objects (e.g., two paintings) placed on the same or adjacent walls.
Within each configuration, we include both \emph{identical} and \emph{unequal} pairs. For identical pairs, the two objects are instantiated from the same 3D asset and therefore have the same physical size. For unequal pairs, we deliberately select two different assets from the same semantic category whose physical dimensions differ along the queried axis. Examples are shown in~\Cref{fig: perspective_scene}.

\noindent\textbf{Perspective camera placement.}
As shown in~\Cref{fig:answerability_overview}(a) right, for each object pair, we generate a \textit{reference view} and a \textit{perspective view}.
The reference view places the camera at an approximately equidistant position from the two objects.
The perspective view shifts the camera toward one object while keeping both objects within the field of view.
The resulting effect depends on the pair configuration.
For \textit{identical-size} pairs, these effects can make physically identical objects appear different in size.
For \textit{unequal-size} pairs, they can reverse the true physical size ordering, such that the physically smaller object appears larger in the image.
Importantly, such perspective-induced disagreement does not necessarily make the underlying physical relation fundamentally unrecoverable from a single image.
For example, objects lying on the same plane may provide projective cues that support geometric recovery~\citep{Criminisi2000SingleVM}.
We therefore distinguish the physical 3D relation from the relation suggested by the 2D projection, and analyze whether model predictions are consistent with the physical relation, follow the projected appearance, or abstain.

\noindent\textbf{Quantifying perspective ambiguity.}
For each object pair, we obtain metric object dimensions from the underlying
3D asset meshes and analytically project the compared dimension into the image
plane using the renderer's pinhole camera model. Let
$\ell_{\mathrm{2D}}(O)$ denote the projected pixel length of the queried
dimension (horizontal width or vertical height) for object $O$.
For unequal-size pairs, we define the projected-size inversion ratio as
$
r_{\mathrm{proj}}
=
\frac{
\ell_{\mathrm{2D}}(O_{\mathrm{small}})
}{
\ell_{\mathrm{2D}}(O_{\mathrm{large}})
},$
where $O_{\mathrm{small}}$ and $O_{\mathrm{large}}$ are determined from the
physical 3D size of the corresponding object dimension.
A ratio above $1$ indicates that the physically smaller object appears larger
in the image, yielding a reversal between the projected 2D relation and the physical 3D relation. Larger values indicate stronger 2D-3D conflict.

\subsection{Task Design}
\label{question design}

\noindent\textbf{Spatial question design.}
For occlusion, we consider four question types: \emph{Visibility}, \emph{Relative position}, \emph{Depth ordering}, and \emph{Size comparison}.
Answerability is defined separately for each type. 
For partially occluded observations, partial visibility alone does not guarantee that a question remains answerable. We therefore retain only questions for which human reviewers can reliably recover the geometry-derived answer from the visible evidence.
For fully occluded observations, visibility questions remain answerable, while questions requiring access to the hidden target are treated as unanswerable, with \emph{Cannot determine} as the appropriate response.
For the perspective setting, we focus specifically on physical size comparison.
For each object pair, we query either the horizontal or vertical dimension.
We define a \emph{2D-3D conflict} when the size relation suggested by the projected 2D appearance disagrees with this physical 3D relation.
In such cases, we characterize model predictions as \emph{3D-consistent}, \emph{2D Appearance-following}, \emph{Abstain}.
For 2D-3D conflict cases, we treat 3D-consistent predictions as correct, 2D-following predictions as errors induced by misleading appearance, and abstention as a conservative response that is reported separately rather
than as correctness.

\noindent\textbf{View selection under perspective ambiguity.}
A reliable model should also recognize when additional visual evidence would support a more
reliable judgment.
Therefore, we introduce two complementary tasks for viewpoint assessment under perspective ambiguity.
\textbf{ViewSelect (ViewSel)}: the model is presented with five candidate views (one informative reference view and four ambiguous alternatives) and asked to identify the view that best supports answering a spatial reasoning question about physical size. This metric evaluates pure viewpoint selection ability in isolation, independent of abstention behavior.
\textbf{Abstain-then-ViewSelect (AbstainViewSel)}: We introduce a two-stage evaluation that tests whether a model can either reason correctly from the current perspective-conflict view or recognize when an additional view is needed. In Stage 1, a 3D-consistent answer is counted as correct directly. If the model abstains, Stage 2 presents five candidate views and asks it to select the informative reference view. The prediction is counted as correct if either the Stage-1 answer is 3D-consistent or the model abstains and successfully selects the informative view.

\subsection{Dataset Statistics and Human Validation}
\label{quality control}

We collect 240 unique scenes spanning 43 room types and construct 1,222 occlusion units, including 618 partial- and 434 full-occlusion cases, yielding 5928 QA pairs across four spatial question types. For each unit, we additionally quantify occlusion severity using the target visible fraction computed from paired instance-mask renderings with and without the designated occluder. For perspective ambiguity, we focus on physical size comparison across four configurations: 150 floor-identical, 186 wall-identical, 83 floor-unequal, and 209 wall-unequal pairs. Each configuration includes a perspective-shifted view designed to create disagreement between projected 2D appearance and the underlying 3D size relation. In total, the perspective set contains 627 2D-3D conflict questions and 718 clean-view controls. More statistics and examples are shown in~\Cref{fig:stats}(b).

All automatically generated configurations undergo human verification for quality control. Evaluation labels are derived from the underlying 3D geometry and camera projection rather than assigned by annotators. For occlusion, reviewers verify rendering quality and, for partially occluded views, additionally filter questions whose geometry-derived answers cannot be reliably recovered from the current observation. Occlusion severity is defined by the measured target visible fraction \(r_{\mathrm{vis}}\), with full occlusion corresponding to less than 5\% visibility. For perspective ambiguity, reviewers verify object identifiability and rendering quality, while 2D-3D conflict labels are computed automatically from the projected-size ratio \(r_{\mathrm{proj}}\). Full annotation protocols and filtering statistics are provided in~\Cref{human_anno_appendix} in the Appendix.

\section{Experimental Results}
\label{main: exp}

\begin{table}[t]
\centering
\small
\caption{
Performance under occlusion and perspective ambiguity.
\textbf{Ans.} and \textbf{Unans.} denote accuracy on answerable and unanswerable cases.
For perspective conflict, \textbf{3D-Cons.}, \textbf{Abstain}, and \textbf{2D-Foll.} denote 3D-consistent, abstaining, and 2D-following predictions.
\textbf{ViewS.} and \textbf{AbsViewS.} denote ViewSel and AbstainViewSel accuracy.
}
\setlength{\tabcolsep}{3.5pt}
\resizebox{\linewidth}{!}{
\begin{tabular}{lcc|c|ccc|cc}
\toprule
\textbf{Model}
& \multicolumn{2}{c}{\textbf{Occlusion}}
& \multicolumn{6}{c}{\textbf{Perspective Ambiguity}} \\
\cmidrule(lr){2-3}
\cmidrule(lr){4-9}

& \multirow{2}{*}{\textbf{Ans.}$\uparrow$}
& \multirow{2}{*}{\textbf{Unans.}$\uparrow$}
& \multirow{2}{*}{\textbf{Ans.}$\uparrow$}
& \multicolumn{3}{c}{\textbf{2D--3D Conflict}}
& \multirow{2}{*}{\textbf{ViewS}$\uparrow$}
& \multirow{2}{*}{\textbf{AbsViewS}$\uparrow$} \\
\cmidrule(lr){5-7}

&
&
&
&
\textbf{3D-Cons.}$\uparrow$
& \textbf{Abstain}
& \textbf{2D-Foll.}$\downarrow$
&
& \\
\midrule

Random
& 32.3 & 23.3
& 25.0
& 25.0 & 25.0 & 25.0
& 20.0 & 30.0 \\
\midrule

\rowcolor{gray!15}
\multicolumn{9}{l}{\textit{Open-source}} \\

Qwen2.5-VL-7B~\citep{Bai2025Qwen25VLTR}
& 51.6 & 42.0
& 19.9
& 3.8 & 27.1 & 65.6
& 24.6 & 5.6 \\

Qwen2.5-VL-32B~\citep{Bai2025Qwen25VLTR}
& 52.0 & 45.5
& 65.7
& 22.8 & 16.4 & 55.2
& 20.7 & 3.8 \\

InternVL3-8B~\citep{Zhu2025InternVL3EA}
& 62.9 & 8.3
& 68.8
& 29.0 & 0.8 & 65.7
& 18.5 & 0.0 \\

\rowcolor{gray!15}
\multicolumn{9}{l}{\textit{Closed-source}} \\

GPT-4o~\citep{gpt4o}
& 43.7 & \textbf{63.6}
& 34.4
& 15.0 & \textbf{38.3} & \textbf{35.6}
& 39.3 & 26.0 \\

GPT-5-mini~\citep{singh2025openai}
& \textbf{66.2} & 8.4
& \textbf{70.8}
& 26.6 & 8.9 & 63.5
& 53.7 & \textbf{34.1} \\

GPT-5.4~\citep{OpenAI2026GPT54}
& 59.0 & 21.1
& 65.7
& 15.6 & 16.7 & 61.1
& \textbf{70.9} & 28.5 \\

Gemini-2.5-Flash~\citep{gemini2.5flash2025}
& 56.8 & 49.4
& 67.4
& \textbf{32.5} & 4.8 & 57.7
& 18.5 & 33.8 \\

Gemini-3.0-Flash~\citep{gemini3flash2025}
& 63.3 & 51.4
& 65.9
& 22.6 & 6.2 & 68.4
& 50.3 & 24.8 \\

\bottomrule
\end{tabular}
}
\label{tab:main_results}
\vspace{-4mm}
\end{table}

\subsection{Evaluation Models and Protocol}

We evaluate eight vision-language models spanning both open- and closed-source families. Open-source models include Qwen2.5-VL-7B~\citep{Bai2025Qwen25VLTR}, Qwen2.5-VL-32B~\citep{Bai2025Qwen25VLTR}, and InternVL3-8B~\citep{Zhu2025InternVL3EA}. Closed-source models include GPT-4o~\citep{gpt4o}, GPT-5-mini~\citep{singh2025openai}, GPT-5.4~\citep{OpenAI2026GPT54}, Gemini-2.5-Flash~\citep{gemini2.5flash2025}, and Gemini-3.0-Flash~\citep{gemini3flash2025}. All models are evaluated zero-shot using standardized multiple-choice prompts (Appendix~\ref{app:prompt}).
For occlusion, models are evaluated on clean, partially occluded, and fully occluded observations. We report \textbf{Ans.}, accuracy on answerable questions, and \textbf{Unans.}, accuracy on cases where the appropriate response is \textit{Cannot determine}. For perspective ambiguity, we separately characterize predictions on 2D-3D conflict cases as \textbf{3D-Cons.}, \textbf{2D-Foll.}, \textbf{Abstain}, depending on whether the prediction follows the physical 3D relation, the projected 2D appearance, or abstains. For viewpoint selection, we report \textbf{ViewSel} and \textbf{AbstainViewSel} accuracy. 
Additional metric details are provided in Appendix~\ref{app: evaluation metrics}. To rule out answer-format artifacts, we shuffle answer options, include a text-only baseline, and additionally verify on 600 open-ended questions that performance closely matches the multiple-choice setting (Appendix~\ref{app:open_ended}).

\begin{figure}
    \centering
    \includegraphics[width=0.95\linewidth]{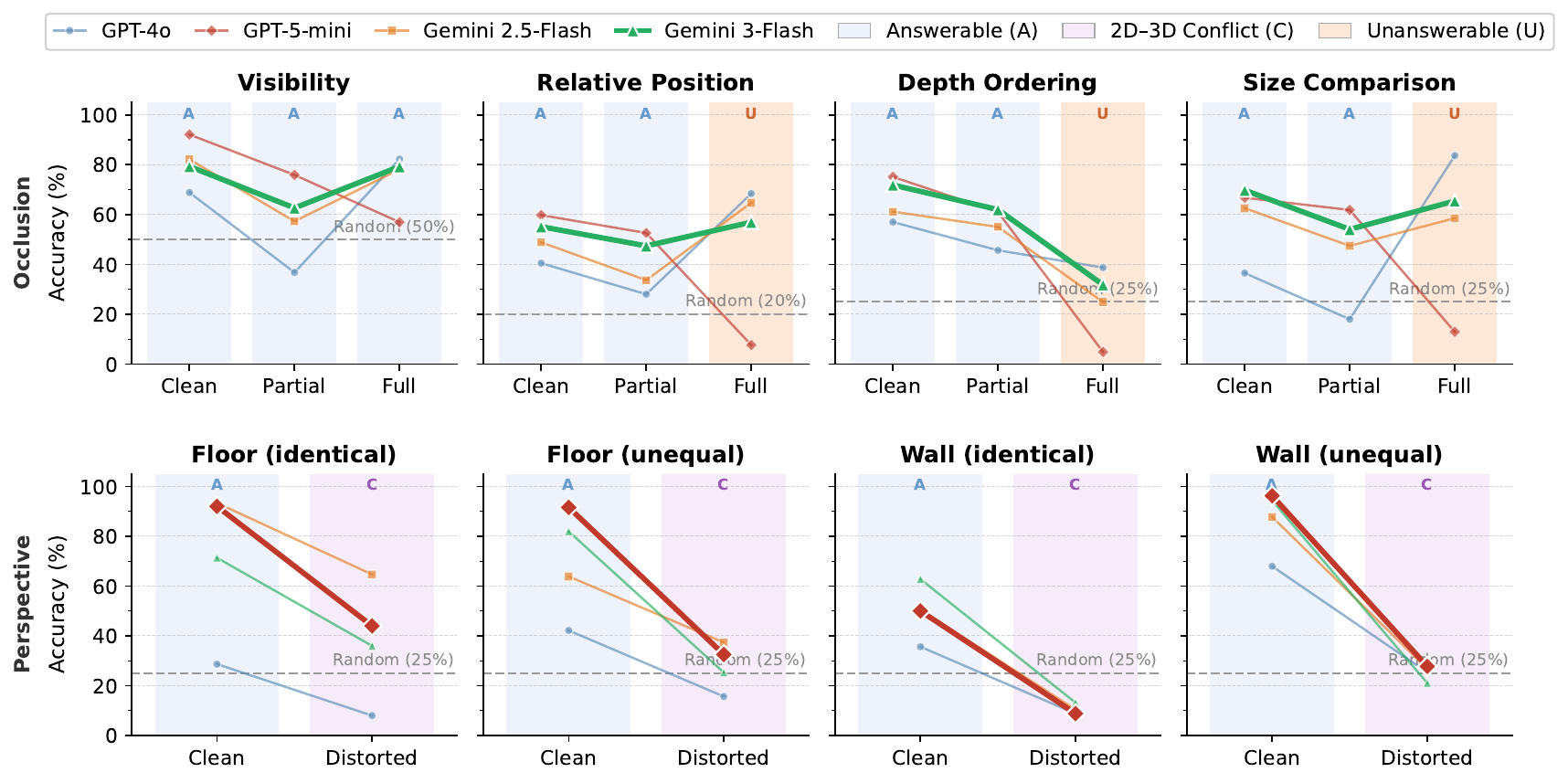}
    \caption{
Model accuracy across question types under occlusion (top) and perspective ambiguity (bottom).
Shaded regions indicate answerable (A), unanswerable (U), and 2D-3D conflict (C) conditions, with dashed lines showing random baselines.
Bold lines highlight Gemini-3.0-Flash for occlusion and GPT-5-mini for perspective ambiguity.
}
    \label{fig:task_breakdown}
    \vspace{-5mm}
\end{figure}

\subsection{Model Behavior under Observational Uncertainty}
\label{sec:main result}

\noindent\textbf{Failure to handle unreliable observations.}
As shown in~\Cref{tab:main_results}, across nearly all models, performance on answerable questions exceeds random chance, indicating meaningful spatial reasoning when visual evidence is sufficient. However, model behavior changes substantially when observations become unreliable, revealing three recurring patterns.
\noindent\textbf{(1) Answer-abstention trade-off.}
Models that abstain more readily often sacrifice accuracy on answerable questions. For example, Gemini-2.5-Flash achieves relatively high Occ-Unans. (49.4) but lower Occ-Ans. (56.8), while GPT-5-mini achieves the highest perspective Ans. (70.8) but abstains on only 8.9 of 2D--3D conflict cases. GPT-4o illustrates the opposite pattern, achieving the highest Occ-Unans. (63.6) but the lowest Occ-Ans. among evaluated models (43.7). These results suggest that current models often improve abstention by becoming more conservative overall, rather than selectively recognizing unreliable evidence.
\noindent\textbf{(2) Inconsistent uncertainty behavior across observation types.}
The same model can respond very differently to different forms of unreliable evidence. For example, Gemini-2.5-Flash achieves Occ-Unans. 49.4 under occlusion but abstains on only 4.8 of perspective-conflict cases, while GPT-4o achieves Occ-Unans. 63.6 and an Abstain rate of 38.3 under perspective conflict. Such variation suggests that abstention behavior does not transfer consistently across different sources of observational uncertainty.
\noindent\textbf{(3) Sensitivity to observation reliability.}
As shown in~\Cref{fig:task_breakdown}, performance declines as occlusion increases. Partial occlusion already hurts performance, while full occlusion reveals models' limited ability to recognize when the queried relation is no longer visually determined.
Under perspective ambiguity, models answer clean reference views reliably, but consistently favor the misleading projection on 2D-3D conflict views. Together, these results show that models do not reliably adjust their behavior when visual evidence becomes incomplete or misleading.

\noindent\textbf{Failure to identify informative viewpoints.}
Results on view selection tasks reveal that current models struggle not only to recognize unreliable observations, but also to identify which additional viewpoints would provide reliable evidence. On \textit{ViewSel}, stronger models such as GPT-5.4 (70.9) and GPT-5-mini (53.7) achieve substantially above-random performance, indicating that models can often identify informative views when explicitly prompted. However, performance remains substantially lower on \textit{AbstainViewSel}, which requires models to either produce a 3D-consistent answer from the current view or abstain and select a more informative viewpoint. This gap suggests that models can often identify an informative view when explicitly prompted, but still struggle to determine whether the current observation is reliable enough to answer directly or whether additional evidence is needed.

\subsection{Diagnosing Observational Failures}
\label{sec:why}

\begin{figure*}[t]
    \centering

    \begin{minipage}[t]{0.48\textwidth}
        \centering
        \includegraphics[width=\linewidth]{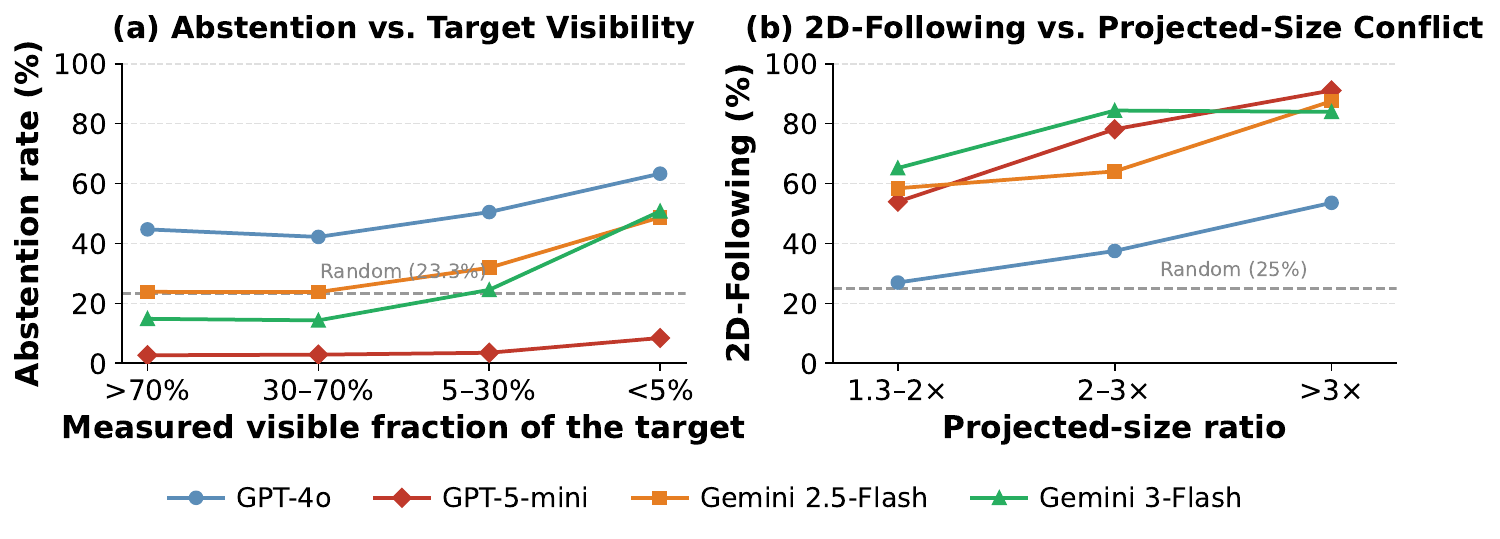}
        \captionof{figure}{
        Effect of observation severity on model behavior.
        \textbf{(a)} Abstention rate under occlusion.
        \textbf{(b)} 2D-following rate with perspective shifts.
        }
        \label{fig:severity}
    \end{minipage}
    \hfill
    \begin{minipage}[t]{0.48\textwidth}
        \centering
        \includegraphics[width=\linewidth]{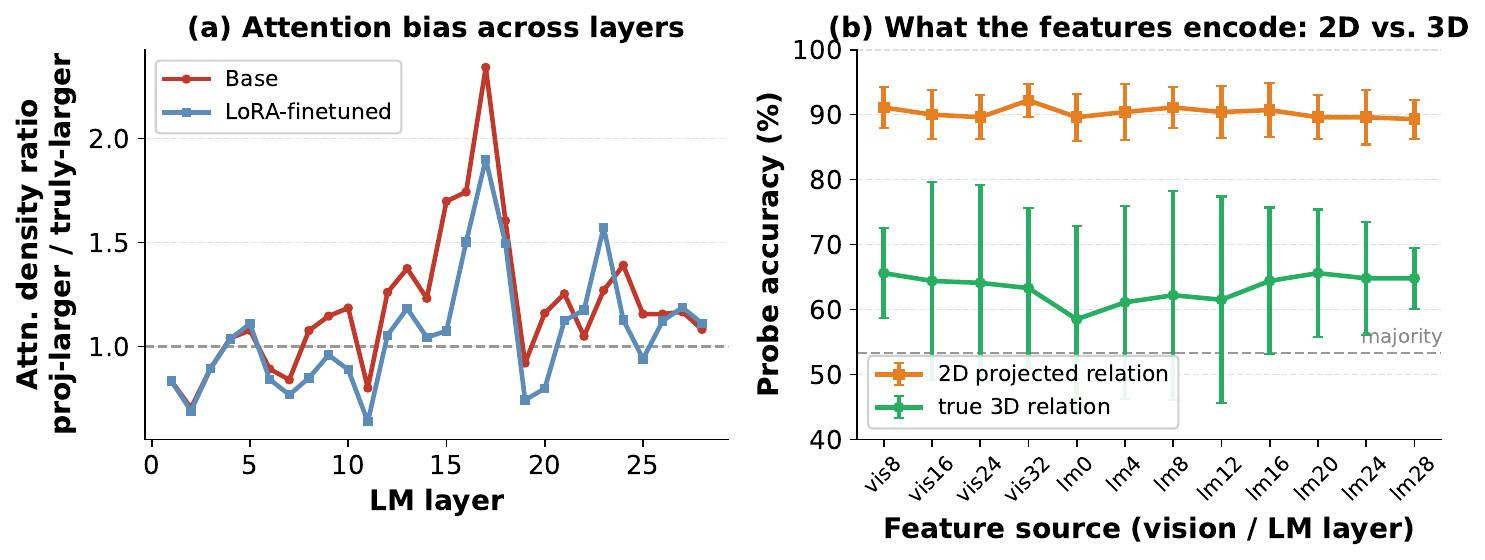}
        \captionof{figure}{
       Internal analysis of projected 2D vs.\ physical 3D cues on Qwen2.5-VL-7B.
        }
        \label{fig:internal}
    \end{minipage}
\vspace{-5mm}
\end{figure*}

\noindent\textbf{Occlusion: abstention only weakly tracks missing evidence.}
Models perform well on visibility questions, suggesting that they can recognize when the target becomes occluded. Their abstention rates also tend to increase as the measured visible fraction decreases, but the response is weak and highly model-dependent (\Cref{fig:severity}a). Even when less than 5\% of the target remains visible, some models still rarely abstain. These results suggest that models can perceive degraded visual evidence, but do not consistently translate its severity into an appropriate decision to abstain.

\noindent\textbf{Perspective: projected 2D cues provide a stronger internal signal.}
To isolate the source of perspective failures, we exploit our controlled construction to vary projected appearance while keeping the underlying physical 3D relation fixed. As the projected-size conflict becomes stronger, models increasingly follow the projected 2D relation (\Cref{fig:severity}b), showing that their predictions systematically shift with viewpoint-dependent appearance rather than the unchanged physical relation. We further examine this behavior internally using Qwen2.5-VL-7B. While the projected 2D relation is readily decodable across model layers, the underlying physical 3D relation remains only weakly decodable and close to chance (\Cref{fig:internal}b), suggesting that the model lacks a robust internal signal for recovering physical size under misleading projections. Answer-token attention is also biased toward the object that appears larger in projection (\Cref{fig:internal}a). Interestingly, this attention bias is reduced after LoRA fine-tuning on our mixed occlusion and perspective training data, consistent with the substantially lower 2D-following behavior after fine-tuning. Together, these results suggest that when the physical 3D relation is weakly represented, models tend to rely on the more accessible projected appearance. 
The same behavioral pattern persists under open-ended evaluation (Appendix~\ref{app:open_ended} and~\Cref{fig:real_image}) and transfers to real images.
Among 35 real-image object pairs that do not appear to belong to a matched set, GPT-5-mini gives contradictory size judgments across viewpoints on 30 pairs (86\%), and Gemini-2.5-Flash on 23 pairs (66\%), each time favoring the object that projects larger. Because physical size is viewpoint-invariant, these contradictions provide a label-free certificate that at least one prediction is incorrect, showing that the effect is not specific to synthetic rendering.

\subsection{Toward Improvement: What Helps and What Does Not}
\label{sec:improve}

\noindent\textbf{Prompting and fine-tuning.}
We first test whether simple interventions can mitigate these failure modes.
As shown in~\Cref{app:prompt} and ~\Cref{tab:prompting} in the Appendix, a structured prompt that explicitly asks models to assess target visibility and viewpoint reliability increases abstention under
occlusion, but at the cost of answerable accuracy. For GPT-5-mini,
Occ-Unans.\ increases from 8.4 to 30.4, while Occ-Ans.\ decreases from
66.2 to 54.7. In contrast, LoRA fine-tuning, shown in~\Cref{tab:lora}, produces broader gains.
For Qwen2.5-VL-7B, it substantially improves both answerable and unanswerable performance under occlusion and nearly eliminates
2D-following under perspective conflict. However, most perspective
predictions shift toward abstention (86.2) rather than 3D-consistent recovery (13.8). These results suggest that prompting primarily encourages more conservative behavior, while supervision can more effectively teach models to recognize when visual cues are unreliable.

\begin{wrapfigure}{r}{0.45\textwidth}
    \centering
     \vspace{-5mm}
    \includegraphics[width=\linewidth]{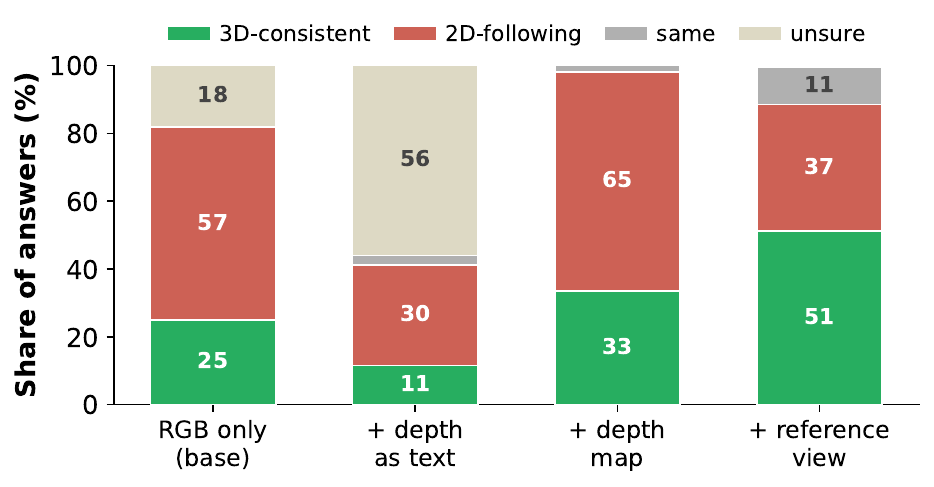}
    \caption{Effect of additional visual evidence on perspective reasoning.}
    \vspace{-3mm}
    \label{fig:intervention_result}
\end{wrapfigure}

\noindent\textbf{Changing the viewpoint is more effective than adding depth information.}
To test what evidence can mitigate perspective-induced errors, we compare three inference-time interventions: ground-truth depth provided as text, a pixel-aligned depth map, and an alternative reference view. As shown in~\Cref{fig:intervention_result}, depth information does not reliably resolve the 2D--3D conflict. Depth as text largely shifts predictions toward uncertainty, while the depth map increases both 3D-consistent predictions (25 to 33) and 2D-following predictions (57 to 65), indicating that it does not consistently steer the model toward the underlying physical relation. In contrast, an alternative viewpoint increases 3D-consistent predictions from 25 to 51 while reducing 2D-following. This suggests that changing the observation can be more effective than augmenting a misleading view with depth information. These results motivate an active-perception perspective: models should recognize when the current observation is unreliable and seek a more informative viewpoint rather than relying on additional cues from the same view.

\section{Conclusion}
We present \methodName, a controlled framework for evaluating whether VLMs can assess the reliability of visual observations under occlusion and perspective ambiguity. Across eight models, behavior often fails to track observation reliability, with models remaining insufficiently cautious under missing evidence and overly influenced by projected appearance. Fine-tuning reduces this reliance largely by increasing abstention rather than recovering the underlying 3D relation, motivating future work on spatial reasoning that can recognize unreliable observations and seek better evidence.

\section{Acknowledgement}
We would like to thank Zengqi Zhao for his help
in the human verification process. 
This work was supported by NSF-AI Engage Institute DRL-2112635, ARO Award W911NF2110220, and ONR Grant N00014-23-1-2356. The views contained in this article are those of the authors and not of the funding agency.

\bibliography{iclr2027_conference}

@inproceedings{yang2025thinking,
  title={Thinking in space: How multimodal large language models see, remember, and recall spaces},
  author={Yang, Jihan and Yang, Shusheng and Gupta, Anjali W and Han, Rilyn and Fei-Fei, Li and Xie, Saining},
  booktitle={Proceedings of the Computer Vision and Pattern Recognition Conference},
  pages={10632--10643},
  year={2025}
}

@inproceedings{Zhang2023VLNTransTF,
  title={VLN-Trans: Translator for the Vision and Language Navigation Agent},
  author={Yue Zhang and Parisa Kordjamshidi},
  booktitle={Annual Meeting of the Association for Computational Linguistics},
  year={2023},
  url={https://api.semanticscholar.org/CorpusID:257038436}
}

@article{Yu2026WhenAH,
  title={When and How Much to Imagine: Adaptive Test-Time Scaling with World Models for Visual Spatial Reasoning},
  author={Shoubin Yu and Yue Zhang and Zun Wang and Jaehong Yoon and Huaxiu Yao and Mingyu Ding and Mohit Bansal},
  journal={ArXiv},
  year={2026},
  volume={abs/2602.08236},
  url={https://api.semanticscholar.org/CorpusID:285452504}
}

@article{zhang2024spartun3d,
  title={Spartun3d: Situated spatial understanding of 3d world in large language models},
  author={Zhang, Yue and Xu, Zhiyang and Shen, Ying and Kordjamshidi, Parisa and Huang, Lifu},
  journal={arXiv preprint arXiv:2410.03878},
  year={2024}
}

@inproceedings{yang2026cambrian,
  title={Cambrian-s: Towards spatial supersensing in video},
  author={Yang, Shusheng and Yang, Jihan and Huang, Pinzhi and Brown, Ellis and Yang, Zihao and Yu, Yue and Tong, Shengbang and Zheng, Zihan and Xu, Yifan and Wang, Muhan and others},
  booktitle={International Conference on Learning Representations},
  volume={2026},
  pages={78185--78225},
  year={2026}
}

@article{zhang2024vision,
  title={Vision-and-language navigation today and tomorrow: A survey in the era of foundation models},
  author={Zhang, Yue and Ma, Ziqiao and Li, Jialu and Qiao, Yanyuan and Wang, Zun and Chai, Joyce and Wu, Qi and Bansal, Mohit and Kordjamshidi, Parisa},
  journal={arXiv preprint arXiv:2407.07035},
  year={2024}
}

@article{liu2023visual,
  title={Visual instruction tuning},
  author={Liu, Haotian and Li, Chunyuan and Wu, Qingyang and Lee, Yong Jae},
  journal={Advances in neural information processing systems},
  volume={36},
  pages={34892--34916},
  year={2023}
}

@article{ma2022sqa3d,
  title={Sqa3d: Situated question answering in 3d scenes},
  author={Ma, Xiaojian and Yong, Silong and Zheng, Zilong and Li, Qing and Liang, Yitao and Zhu, Song-Chun and Huang, Siyuan},
  journal={arXiv preprint arXiv:2210.07474},
  year={2022}
}

@inproceedings{daxberger2025mm,
  title={Mm-spatial: Exploring 3d spatial understanding in multimodal llms},
  author={Daxberger, Erik and Wenzel, Nina and Griffiths, David and Gang, Haiming and Lazarow, Justin and Kohavi, Gefen and Kang, Kai and Eichner, Marcin and Yang, Yinfei and Dehghan, Afshin and others},
  booktitle={Proceedings of the IEEE/CVF International Conference on Computer Vision},
  pages={7395--7408},
  year={2025}
}

@inproceedings{wang2024embodiedscan,
  title={Embodiedscan: A holistic multi-modal 3d perception suite towards embodied ai},
  author={Wang, Tai and Mao, Xiaohan and Zhu, Chenming and Xu, Runsen and Lyu, Ruiyuan and Li, Peisen and Chen, Xiao and Zhang, Wenwei and Chen, Kai and Xue, Tianfan and others},
  booktitle={Proceedings of the IEEE/CVF Conference on Computer Vision and Pattern Recognition},
  pages={19757--19767},
  year={2024}
}

@article{Rajabi2024GSRBENCHAB,
  title={GSR-BENCH: A Benchmark for Grounded Spatial Reasoning Evaluation via Multimodal LLMs},
  author={Navid Rajabi and Jana Kosecka},
  journal={ArXiv},
  year={2024},
  volume={abs/2406.13246},
  url={https://api.semanticscholar.org/CorpusID:270619607}
}

@inproceedings{johnson2017clevr,
  title={Clevr: A diagnostic dataset for compositional language and elementary visual reasoning},
  author={Johnson, Justin and Hariharan, Bharath and Van Der Maaten, Laurens and Fei-Fei, Li and Lawrence Zitnick, C and Girshick, Ross},
  booktitle={Proceedings of the IEEE conference on computer vision and pattern recognition},
  pages={2901--2910},
  year={2017}
}

@article{Cheng2024SpatialRGPTGS,
  title={SpatialRGPT: Grounded Spatial Reasoning in Vision Language Model},
  author={An-Chieh Cheng and Hongxu Yin and Yang Fu and Qiushan Guo and Ruihan Yang and Jan Kautz and Xiaolong Wang and Sifei Liu},
  journal={ArXiv},
  year={2024},
  volume={abs/2406.01584},
  url={https://api.semanticscholar.org/CorpusID:270215984}
}

@inproceedings{chen2024spatialvlm,
  title={Spatialvlm: Endowing vision-language models with spatial reasoning capabilities},
  author={Chen, Boyuan and Xu, Zhuo and Kirmani, Sean and Ichter, Brain and Sadigh, Dorsa and Guibas, Leonidas and Xia, Fei},
  booktitle={Proceedings of the IEEE/CVF Conference on Computer Vision and Pattern Recognition},
  pages={14455--14465},
  year={2024}
}

@article{xu2025spatialbench,
  title={Spatialbench: Benchmarking multimodal large language models for spatial cognition},
  author={Xu, Peiran and Wang, Sudong and Zhu, Yao and Li, Jianing and Qi, Gege and Zhang, Yunjian},
  journal={arXiv preprint arXiv:2511.21471},
  year={2025}
}

@inproceedings{pothiraj2025capture,
  title={Capture: Evaluating spatial reasoning in vision language models via occluded object counting},
  author={Pothiraj, Atin and Stengel-Eskin, Elias and Cho, Jaemin and Bansal, Mohit},
  booktitle={Proceedings of the IEEE/CVF International Conference on Computer Vision},
  pages={8001--8010},
  year={2025}
}

@article{wang2024picture,
  title={Is a picture worth a thousand words? delving into spatial reasoning for vision language models},
  author={Wang, Jiayu and Ming, Yifei and Shi, Zhenmei and Vineet, Vibhav and Wang, Xin and Li, Yixuan and Joshi, Neel},
  journal={Advances in Neural Information Processing Systems},
  volume={37},
  pages={75392--75421},
  year={2024}
}

@inproceedings{liu2025can,
  title={Can Multimodal Large Language Models Understand Spatial Relations?},
  author={Liu, Jingping and Liu, Ziyan and Cen, Zhedong and Zhou, Yan and Zou, Yinan and Zhang, Weiyan and Jiang, Haiyun and Ruan, Tong},
  booktitle={Proceedings of the 63rd Annual Meeting of the Association for Computational Linguistics (Volume 1: Long Papers)},
  pages={620--632},
  year={2025}
}

@article{yang2025mmsi,
  title={Mmsi-bench: A benchmark for multi-image spatial intelligence},
  author={Yang, Sihan and Xu, Runsen and Xie, Yiman and Yang, Sizhe and Li, Mo and Lin, Jingli and Zhu, Chenming and Chen, Xiaochen and Duan, Haodong and Yue, Xiangyu and others},
  journal={arXiv preprint arXiv:2505.23764},
  year={2025}
}

@article{jia2025omnispatial,
  title={Omnispatial: Towards comprehensive spatial reasoning benchmark for vision language models},
  author={Jia, Mengdi and Qi, Zekun and Zhang, Shaochen and Zhang, Wenyao and Yu, Xinqiang and He, Jiawei and Wang, He and Yi, Li},
  journal={arXiv preprint arXiv:2506.03135},
  year={2025}
}

@article{stogiannidis2025mind,
  title={Mind the gap: Benchmarking spatial reasoning in vision-language models},
  author={Stogiannidis, Ilias and McDonagh, Steven and Tsaftaris, Sotirios A},
  journal={arXiv preprint arXiv:2503.19707},
  year={2025}
}

@inproceedings{yang2024holodeck,
  title={Holodeck: Language guided generation of 3d embodied ai environments},
  author={Yang, Yue and Sun, Fan-Yun and Weihs, Luca and VanderBilt, Eli and Herrasti, Alvaro and Han, Winson and Wu, Jiajun and Haber, Nick and Krishna, Ranjay and Liu, Lingjie and others},
  booktitle={Proceedings of the IEEE/CVF Conference on Computer Vision and Pattern Recognition},
  pages={16227--16237},
  year={2024}
}

@inproceedings{zhang2026spinbench,
  title={Spinbench: Perspective and rotation as a lens on spatial reasoning in vlms},
  author={Zhang, Yuyou and Corcodel, Radu and Hori, Chiori and Cherian, Anoop and Zhao, Ding},
  booktitle={International Conference on Learning Representations},
  volume={2026},
  pages={70072--70141},
  year={2026}
}

@article{zhang2026multiview,
  title={MultiView-Bench: A Diagnostic Benchmark for World-Centric Multi-View Integration in VLMs},
  author={Zhang, Hantao and Sui, Jinru and Li, Ed and Bergemann, Dirk and Yang, Zhuoran},
  journal={arXiv preprint arXiv:2607.08970},
  year={2026}
}

@article{duan2022survey,
  title={A survey of embodied ai: From simulators to research tasks},
  author={Duan, Jiafei and Yu, Samson and Tan, Hui Li and Zhu, Hongyuan and Tan, Cheston},
  journal={IEEE Transactions on Emerging Topics in Computational Intelligence},
  volume={6},
  number={2},
  pages={230--244},
  year={2022},
  publisher={IEEE}
}

@article{kolve2017ai2,
  title={Ai2-thor: An interactive 3d environment for visual ai},
  author={Kolve, Eric and Mottaghi, Roozbeh and Han, Winson and VanderBilt, Eli and Weihs, Luca and Herrasti, Alvaro and Deitke, Matt and Ehsani, Kiana and Gordon, Daniel and Zhu, Yuke and others},
  journal={arXiv preprint arXiv:1712.05474},
  year={2017}
}

@misc{gemini2.5flash2025,
  author       = {{Deepmind}},
  title        = {Gemini 2.5: Pushing the frontier with advanced reasoning, multimodality, long context, and next generation agentic capabilities},
  year         = {2025},
  howpublished = {\url{https://arxiv.org/abs/2507.06261}},
}

@misc{gemini3flash2025,
  author       = {{Deepmind}},
  title        = {Gemini 3 Flash: Frontier Intelligence Built for Speed},
  year         = {2025},
  howpublished = {\url{https://blog.google/products/gemini/gemini-3-flash/}},
}

@misc{gpt4o,
  author       = {OpenAI},
  title        = {Hello {GPT}-4o},
  year         = {2024},
  url = {https://openai.com/index/hello-gpt-4o}
}

@article{singh2025openai,
  title={Openai gpt-5 system card},
  author={Singh, Aaditya and Fry, Adam and Perelman, Adam and Tart, Adam and Ganesh, Adi and El-Kishky, Ahmed and McLaughlin, Aidan and Low, Aiden and Ostrow, AJ and Ananthram, Akhila and others},
  journal={arXiv preprint arXiv:2601.03267},
  year={2025}
}

@misc{OpenAI2026GPT54,
title={OpenAI: GPT-5.4 model},
author = {OpenAI},
url ={https://developers.openai.com/api/docs/models/gpt-5.4},
year = {2026}
}

@article{Zhu2025InternVL3EA,
  title={Internvl3: Exploring advanced training and test-time recipes for open-source multimodal models},
  author={Zhu, Jinguo and Wang, Weiyun and Chen, Zhe and Liu, Zhaoyang and Ye, Shenglong and Gu, Lixin and Tian, Hao and Duan, Yuchen and Su, Weijie and Shao, Jie and others},
  journal={arXiv preprint arXiv:2504.10479},
  year={2025}
}

@article{Bai2025Qwen25VLTR,
  title={Qwen2.5-VL Technical Report},
  author={Shuai Bai and Keqin Chen and Xuejing Liu and Jialin Wang and Wenbin Ge and Sibo Song and Kai Dang and Peng Wang and Shijie Wang and Jun Tang and Humen Zhong and Yuanzhi Zhu and Mingkun Yang and Zhaohai Li and Jianqiang Wan and Pengfei Wang and Wei Ding and Zheren Fu and Yiheng Xu and Jiabo Ye and Xi Zhang and Tianbao Xie and Zesen Cheng and Hang Zhang and Zhibo Yang and Haiyang Xu and Junyang Lin},
  journal={ArXiv},
  year={2025},
  volume={abs/2502.13923},
  url={https://api.semanticscholar.org/CorpusID:276449796}
}

@article{stengel2024lacie,
  title={LACIE: Listener-aware finetuning for calibration in large language models},
  author={Stengel-Eskin, Elias and Hase, Peter and Bansal, Mohit},
  journal={Advances in Neural Information Processing Systems},
  volume={37},
  pages={43080--43106},
  year={2024}
}

@inproceedings{guo2017calibration,
  title={On calibration of modern neural networks},
  author={Guo, Chuan and Pleiss, Geoff and Sun, Yu and Weinberger, Kilian Q},
  booktitle={International conference on machine learning},
  pages={1321--1330},
  year={2017},
  organization={PMLR}
}

@article{hendrycks2016baseline,
  title={A baseline for detecting misclassified and out-of-distribution examples in neural networks},
  author={Hendrycks, Dan and Gimpel, Kevin},
  journal={arXiv preprint arXiv:1610.02136},
  year={2016}
}

@article{geifman2017selective,
  title={Selective classification for deep neural networks},
  author={Geifman, Yonatan and El-Yaniv, Ran},
  journal={Advances in neural information processing systems},
  volume={30},
  year={2017}
}

@inproceedings{lin2022truthfulqa,
  title={Truthfulqa: Measuring how models mimic human falsehoods},
  author={Lin, Stephanie and Hilton, Jacob and Evans, Owain},
  booktitle={Proceedings of the 60th annual meeting of the association for computational linguistics (volume 1: long papers)},
  pages={3214--3252},
  year={2022}
}

@article{kadavath2022language,
  title={Language models (mostly) know what they know},
  author={Kadavath, Saurav and Conerly, Tom and Askell, Amanda and Henighan, Tom and Drain, Dawn and Perez, Ethan and Schiefer, Nicholas and Hatfield-Dodds, Zac and DasSarma, Nova and Tran-Johnson, Eli and others},
  journal={arXiv preprint arXiv:2207.05221},
  year={2022}
}

@inproceedings{yin2023large,
  title={Do large language models know what they don’t know?},
  author={Yin, Zhangyue and Sun, Qiushi and Guo, Qipeng and Wu, Jiawen and Qiu, Xipeng and Huang, Xuan-Jing},
  booktitle={Findings of the association for Computational Linguistics: ACL 2023},
  pages={8653--8665},
  year={2023}
}

@inproceedings{tian2023just,
  title={Just ask for calibration: Strategies for eliciting calibrated confidence scores from language models fine-tuned with human feedback},
  author={Tian, Katherine and Mitchell, Eric and Zhou, Allan and Sharma, Archit and Rafailov, Rafael and Yao, Huaxiu and Finn, Chelsea and Manning, Christopher D},
  booktitle={Proceedings of the 2023 Conference on Empirical Methods in Natural Language Processing},
  pages={5433--5442},
  year={2023}
}

@inproceedings{
xiong2024can,
title={Can {LLM}s Express Their Uncertainty? An Empirical Evaluation of Confidence Elicitation in {LLM}s},
author={Miao Xiong and Zhiyuan Hu and Xinyang Lu and YIFEI LI and Jie Fu and Junxian He and Bryan Hooi},
booktitle={The Twelfth International Conference on Learning Representations},
year={2024},
url={https://openreview.net/forum?id=gjeQKFxFpZ}
}

@inproceedings{manakul2023selfcheckgpt,
  title={Selfcheckgpt: Zero-resource black-box hallucination detection for generative large language models},
  author={Manakul, Potsawee and Liusie, Adian and Gales, Mark},
  booktitle={Proceedings of the 2023 conference on empirical methods in natural language processing},
  pages={9004--9017},
  year={2023}
}

@article{wen2025know,
  title={Know your limits: A survey of abstention in large language models},
  author={Wen, Bingbing and Yao, Jihan and Feng, Shangbin and Xu, Chenjun and Tsvetkov, Yulia and Howe, Bill and Wang, Lucy Lu},
  journal={Transactions of the Association for Computational Linguistics},
  volume={13},
  pages={529--556},
  year={2025},
  publisher={MIT Press 255 Main Street, 9th Floor, Cambridge, Massachusetts 02142, USA~…}
}

@inproceedings{rohrbach2018object,
  title={Object hallucination in image captioning},
  author={Rohrbach, Anna and Hendricks, Lisa Anne and Burns, Kaylee and Darrell, Trevor and Saenko, Kate},
  booktitle={Proceedings of the 2018 Conference on Empirical Methods in Natural Language Processing},
  pages={4035--4045},
  year={2018}
}

@inproceedings{li2023evaluating,
  title={Evaluating object hallucination in large vision-language models},
  author={Li, Yifan and Du, Yifan and Zhou, Kun and Wang, Jinpeng and Zhao, Xin and Wen, Ji-Rong},
  booktitle={Proceedings of the 2023 conference on empirical methods in natural language processing},
  pages={292--305},
  year={2023}
}

@inproceedings{sun2024aligning,
  title={Aligning large multimodal models with factually augmented rlhf},
  author={Sun, Zhiqing and Shen, Sheng and Cao, Shengcao and Liu, Haotian and Li, Chunyuan and Shen, Yikang and Gan, Chuang and Gui, Liangyan and Wang, Yu-Xiong and Yang, Yiming and others},
  booktitle={Findings of the Association for Computational Linguistics: ACL 2024},
  pages={13088--13110},
  year={2024}
}

@inproceedings{guan2024hallusionbench,
  title={Hallusionbench: an advanced diagnostic suite for entangled language hallucination and visual illusion in large vision-language models},
  author={Guan, Tianrui and Liu, Fuxiao and Wu, Xiyang and Xian, Ruiqi and Li, Zongxia and Liu, Xiaoyu and Wang, Xijun and Chen, Lichang and Huang, Furong and Yacoob, Yaser and others},
  booktitle={Proceedings of the IEEE/CVF conference on computer vision and pattern recognition},
  pages={14375--14385},
  year={2024}
}

@inproceedings{gurari2018vizwiz,
  title={Vizwiz grand challenge: Answering visual questions from blind people},
  author={Gurari, Danna and Li, Qing and Stangl, Abigale J and Guo, Anhong and Lin, Chi and Grauman, Kristen and Luo, Jiebo and Bigham, Jeffrey P},
  booktitle={Proceedings of the IEEE conference on computer vision and pattern recognition},
  pages={3608--3617},
  year={2018}
}

@article{guo2024unk,
  title={Unk-vqa: A dataset and a probe into the abstention ability of multi-modal large models},
  author={Guo, Yangyang and Jiao, Fangkai and Shen, Zhiqi and Nie, Liqiang and Kankanhalli, Mohan},
  journal={IEEE Transactions on Pattern Analysis and Machine Intelligence},
  volume={46},
  number={12},
  pages={10284--10296},
  year={2024},
  publisher={IEEE}
}

@inproceedings{whitehead2022reliable,
  title={Reliable visual question answering: Abstain rather than answer incorrectly},
  author={Whitehead, Spencer and Petryk, Suzanne and Shakib, Vedaad and Gonzalez, Joseph and Darrell, Trevor and Rohrbach, Anna and Rohrbach, Marcus},
  booktitle={European Conference on Computer Vision},
  pages={148--166},
  year={2022},
  organization={Springer}
}

@inproceedings{eisenschlos2024selectively,
  title={Selectively answering visual questions},
  author={Eisenschlos, Julian and Maina, Hern{\'a}n and Ivetta, Guido and Benotti, Luciana},
  booktitle={Findings of the Association for Computational Linguistics: ACL 2024},
  pages={4219--4229},
  year={2024}
}

@article{yu2026and,
  title={When and how much to imagine: Adaptive test-time scaling with world models for visual spatial reasoning},
  author={Yu, Shoubin and Zhang, Yue and Wang, Zun and Yoon, Jaehong and Yao, Huaxiu and Ding, Mingyu and Bansal, Mohit},
  journal={arXiv preprint arXiv:2602.08236},
  year={2026}
}

@inproceedings{kamath2023s,
  title={What’s “up” with vision-language models? investigating their struggle with spatial reasoning},
  author={Kamath, Amita and Hessel, Jack and Chang, Kai-Wei},
  booktitle={Proceedings of the 2023 Conference on Empirical Methods in Natural Language Processing},
  pages={9161--9175},
  year={2023}
}

@article{Criminisi2000SingleVM,
  title={Single View Metrology},
  author={Antonio Criminisi and Ian D. Reid and Andrew Zisserman},
  journal={International Journal of Computer Vision},
  year={2000},
  volume={40},
  pages={123-148},
  url={https://api.semanticscholar.org/CorpusID:2499410}
}

@article{madhusudhan2026knowing,
  title={Knowing When Not to Answer: Evaluating Abstention in Multimodal Reasoning Systems},
  author={Madhusudhan, Nishanth and Yadav, Vikas and Lacoste, Alexandre},
  journal={arXiv preprint arXiv:2604.14799},
  year={2026}
}

@article{kirichenko2026abstentionbench,
  title={Abstentionbench: Reasoning llms fail on unanswerable questions},
  author={Kirichenko, Polina and Ibrahim, Mark and Chaudhuri, Kamalika and Bell, Samuel J},
  journal={Advances in Neural Information Processing Systems},
  volume={38},
  year={2026}
}

@inproceedings{wang2025spatial,
  title={Spatial mental modeling from limited views},
  author={Wang, Qineng and Yin, Baiqiao and Zhang, Pingyue and Zhang, Jianshu and Wang, Kangrui and Wang, Zihan and Zhang, Jieyu and Chandrasegaran, Keshigeyan and Liu, Han and Krishna, Ranjay and others},
  booktitle={The Fourteenth International Conference on Learning Representations},
  year={2025}
}

@article{he2024tubench,
  title={TUBench: Benchmarking large vision-language models on trustworthiness with unanswerable questions},
  author={He, Xingwei and Zhang, Qianru and Jin, A and Yuan, Yuan and Yiu, Siu-Ming and others},
  journal={arXiv preprint arXiv:2410.04107},
  year={2024}
}

@inproceedings{yang2025cambrian,
  title={Cambrian-s: Towards spatial supersensing in video},
  author={Yang, Shusheng and Yang, Jihan and Huang, Pinzhi and Brown II, Ellis L and Yang, Zihao and Yu, Yue and Tong, Shengbang and Zheng, Zihan and Xu, Yifan and Wang, Muhan and others},
  booktitle={The Fourteenth International Conference on Learning Representations},
  year={2025}
}
\bibliographystyle{iclr2027_conference}

\appendix

\section{Appendix}
\subsection{\methodName~Construction Details}
\label{benchmark construction appendix}

\paragraph{Target occluder in occlusion configuration}
We score each object based on its visibility from the camera viewpoint, combining two factors: angular centrality (how close the object is to the camera's optical axis) and apparent size (the object's projected size relative to its depth). Objects that are partially occluded by other scene objects receive a penalty. We retain the top-k (k=3) objects per scene as target candidates, further requiring scene-level uniqueness.

\paragraph{Object pair selection in perspective configuration.}
To induce perspective ambiguity, we identify pairs of objects that are physically comparable but visually sensitive to viewpoint changes. Specifically, we select pairs of same-category objects with similar physical size, ensuring that they are expected to appear comparable under neutral viewpoints but can exhibit large appearance differences under perspective distortion.
We consider two types of object pairs. \textit{Floor pairs} consist of two floor-standing objects of the same type (e.g., chairs), whose centers are below 1.2m and are separated by at least 2.5m to allow for significant depth variation. \textit{Wall pairs} consist of two wall-mounted objects (e.g., paintings) placed on the same or adjacent walls, separated by at least 1.2m, and matched in aspect ratio within a 20\% tolerance to ensure similar physical proportions. 
For each scene, we sample up to 3 floor pairs and 2 wall pairs, promoting diversity while maintaining controlled geometric conditions.

\subsubsection{Human Verification}
\label{human_anno_appendix}
All automatically generated configurations undergo human verification by seven reviewers for quality control. Annotators independently review each configuration, and approximately one third of generated candidates are removed during this process. Evaluation labels themselves are derived from the underlying 3D geometry and camera projection rather than assigned by annotators.

For occlusion, reviewers verify that the designated occluder is correctly rendered and that the resulting scene is visually plausible and free of obvious artifacts. Occlusion severity is determined by the geometrically measured target visible fraction \(r_{\mathrm{vis}}\), computed from paired instance-mask renderings with and without the designated occluder. We define partial occlusion as \(r_{\mathrm{vis}} > 5\%\) and full occlusion as \(r_{\mathrm{vis}} < 5\%\), where targets below this threshold are typically no longer visually identifiable. For partially occluded views, reviewers additionally assess question answerability, and we retain only questions for which the geometry-derived answer can be reliably recovered from the current observation. This step filters ambiguous questions but does not alter their geometry-derived labels.

For perspective ambiguity, reviewers verify that both objects are clearly identifiable and that the reference and perspective views are free of rendering artifacts or ambiguous framing. The 2D--3D conflict labels are computed automatically from the projected-size ratio \(r_{\mathrm{proj}}\). Human review therefore filters configurations and questions for quality and answerability, but does not assign any evaluation label.

\subsection{Evaluation Setup}
\label{app:evaluation}

\subsubsection{Prompt Templates}
\label{app:prompt}
We use two prompt variants in our experiments: a standard multiple-choice prompt and a structured reasoning prompt.

\paragraph{Standard prompt.}
The standard prompt instructs the model to select the best answer based on visible evidence, without any explicit guidance on assessing observation reliability. It permits abstention via ``Cannot determine'' but does not actively encourage it. This prompt serves as our primary evaluation setting.

\paragraph{Structured reasoning prompt.}
The structured reasoning prompt explicitly guides the model to assess observation reliability before selecting an answer. It decomposes the reasoning process into two explicit checks: whether the target is visible, and whether the viewpoint is reliable. Only if both checks pass does the model proceed to select a specific answer; otherwise, it defaults to ``Cannot determine.'' This prompt is used in our prompting analysis to investigate whether explicit reasoning guidance can improve observational awareness.

\begin{tcolorbox}[colback=gray!5,colframe=gray!40,breakable]
\small
You are answering a visual multiple-choice question about the provided image(s).

Look carefully at the image and choose the best answer.

\textbf{Rules:}
\begin{itemize}
    \item You must choose exactly one option.
    \item Choose ``Cannot determine'' if the image lacks sufficient visual evidence to decide reliably.
    \item Reply with ONLY one letter.
\end{itemize}

\textbf{Question:} \\
\{question\}

\textbf{Options:} \\
\{options\}

Reply with ONLY one letter from: A, B, C, \dots
\end{tcolorbox}

\paragraph{Structured Reasoning Prompt.}
\begin{tcolorbox}[colback=gray!5,colframe=gray!40,breakable]
\small
You are answering a visual multiple-choice question about the provided image(s).

Follow these steps before selecting your answer:

\textbf{Step 1:} Is there sufficient visible evidence to answer the question reliably? Answer Yes or No.

\textbf{Step 2:} Could the current viewpoint make the relevant spatial property misleading? Answer Yes or No.

\textbf{Step 3:} If Step 1 is Yes and Step 2 is No, select the correct option. Otherwise, select \{cannot\_letter\}) Cannot determine.

\textbf{Question:} \\
\{question\}

\textbf{Options:} \\
\{options\}

Write your step-by-step reasoning, then on a new line write:

\texttt{Answer: <letter from A, B, C, ...>}
\end{tcolorbox}

\begin{figure}[h]
    \centering
    
    \begin{subfigure}{0.48\linewidth}
        \centering
        \includegraphics[width=\linewidth]{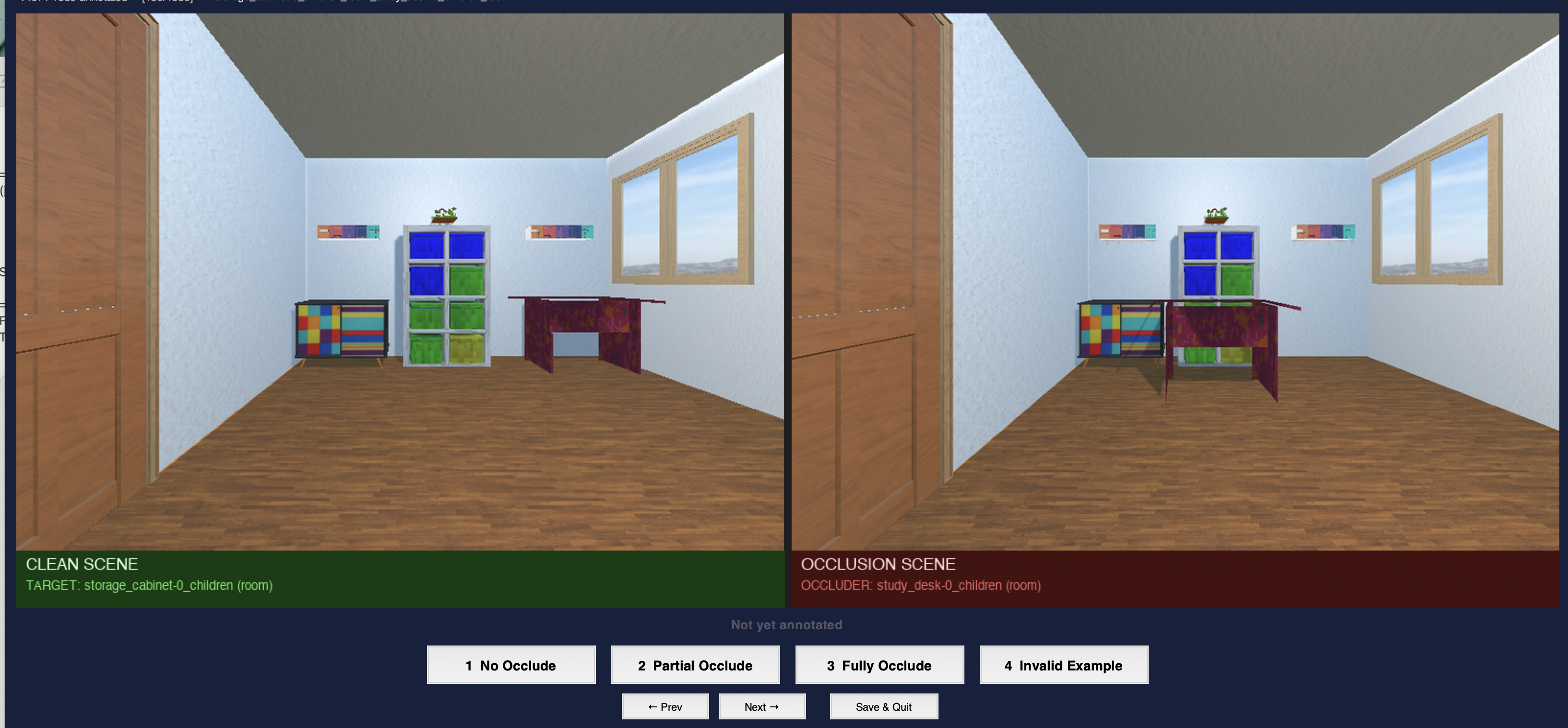}
        \caption{Occlusion annotation examples.}
        \label{fig:occ_anno}
    \end{subfigure}
    \hfill
    \begin{subfigure}{0.48\linewidth}
        \centering
        \includegraphics[width=\linewidth]{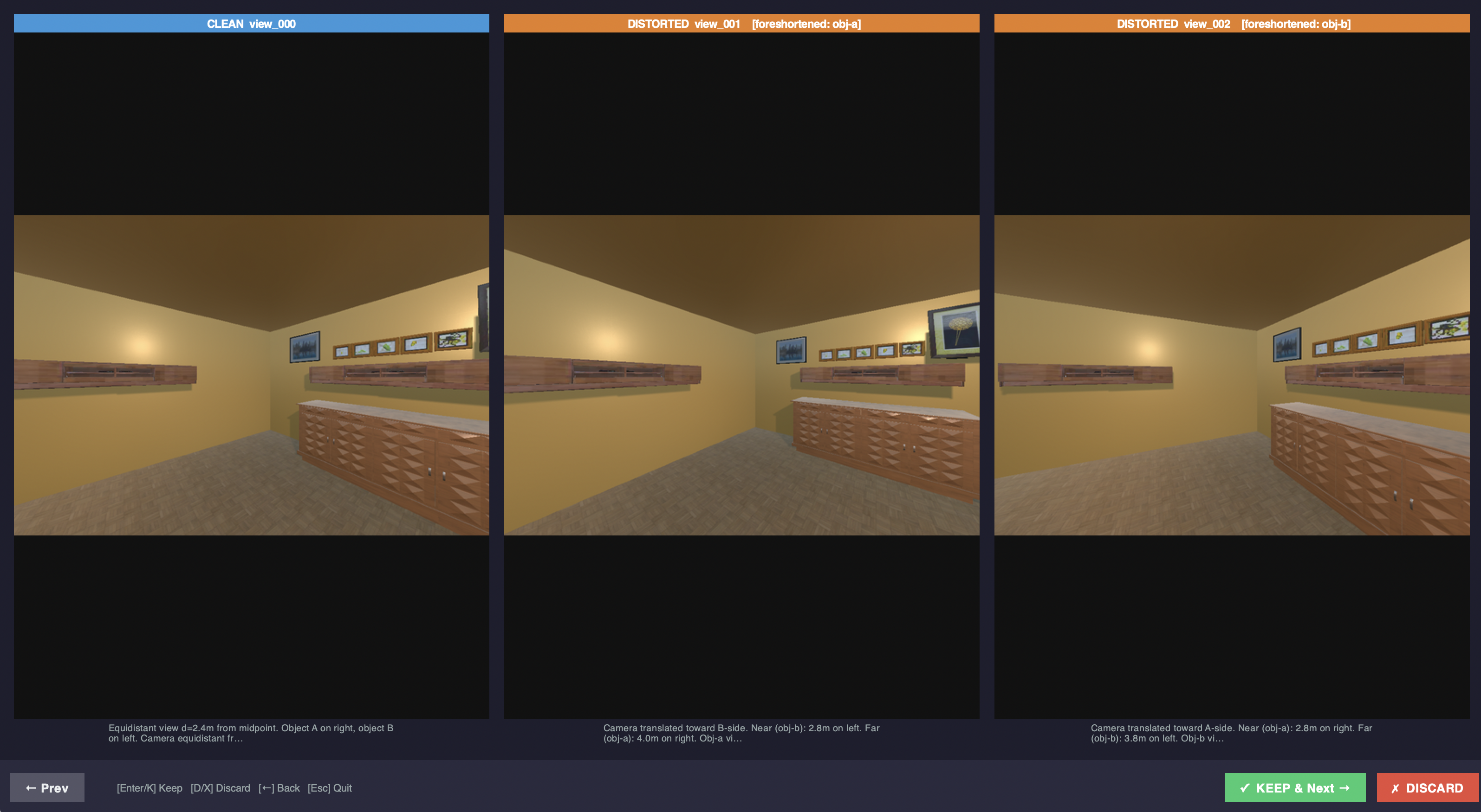}
        \caption{Perspective ambiguity examples.}
        \label{fig:pers_anno}
    \end{subfigure}
    
    \caption{Annotation Interface for Occlusion and Perspective Scenes.}
    \label{fig:annotation_examples}
\end{figure}

\subsubsection{Evaluation Metrics}
\label{app: evaluation metrics}
Models are presented with multiple-choice questions and required to select exactly one option, including \textit{Cannot determine} where applicable. We report the following metrics:
\textbf{Answerable Accuracy (Ans.)} is calculated by 
$
\text{Ans.} =
\frac{\#\text{ correct (ans)}}
{\#\text{ total (ans)}}
$.
\textbf{Unanswerable Accuracy (Unans.)} is calculated by
$
\text{Unans.} =
\frac{\#\text{ correct (unans)}}
{\#\text{ total (unans)}}
$.
\textbf{Overall Accuracy (All)} is calculated by
$
\text{All} =
\frac{
\#\text{ correct (ans)} + \#\text{ correct (unans)}
}{
\#\text{ total (ans)} + \#\text{ total (unans)}
}
$,
\textbf{ViewSel} is calculated by
$
\text{ViewSel} =
\frac{
\#\text{ correctly selected views}
}{
\#\text{ total view selection questions}
}
$, and \textbf{AbstainViewSel} is calculated by 
$
\text{AbstainViewSel} =
\frac{
\#\text{3D-consistent Stage-1 predictions}
+
\#\text{successful abstain-and-select cases}
}{
\#\text{perspective-conflict questions}
}
$.

\subsubsection{Implementation Details}
\label{aa:implementation details}
We fine-tune Qwen2.5-VL-7B-Instruct with LoRA adapters on the occlusion and perspective training splits separately to study cross-condition abstention transfer. LoRA is applied to all linear projections in the language tower ($r{=}16$, $\alpha{=}32$, dropout $0.05$), while the vision encoder remains frozen. Training uses bf16, gradient checkpointing, cosine learning-rate scheduling, and a warm-up ratio of 0.03. Loss is computed only on assistant response tokens.
The occlusion adapter is trained on 5.2K samples for 1 epoch with learning rate $3\mathrm{e}{-5}$, batch size 4, and gradient accumulation 2. The perspective adapter is trained on 3.0K samples for 2 epochs with learning rate $1\mathrm{e}{-4}$, batch size 2, and gradient accumulation 8. Training is conducted on 2$\times$A100 80GB GPUs per adapter. At evaluation, each adapter is tested on held-out scenes from both benchmarks to measure in-domain and cross-domain abstention transfer.

\begin{figure}
    \centering
    \includegraphics[width=\linewidth]{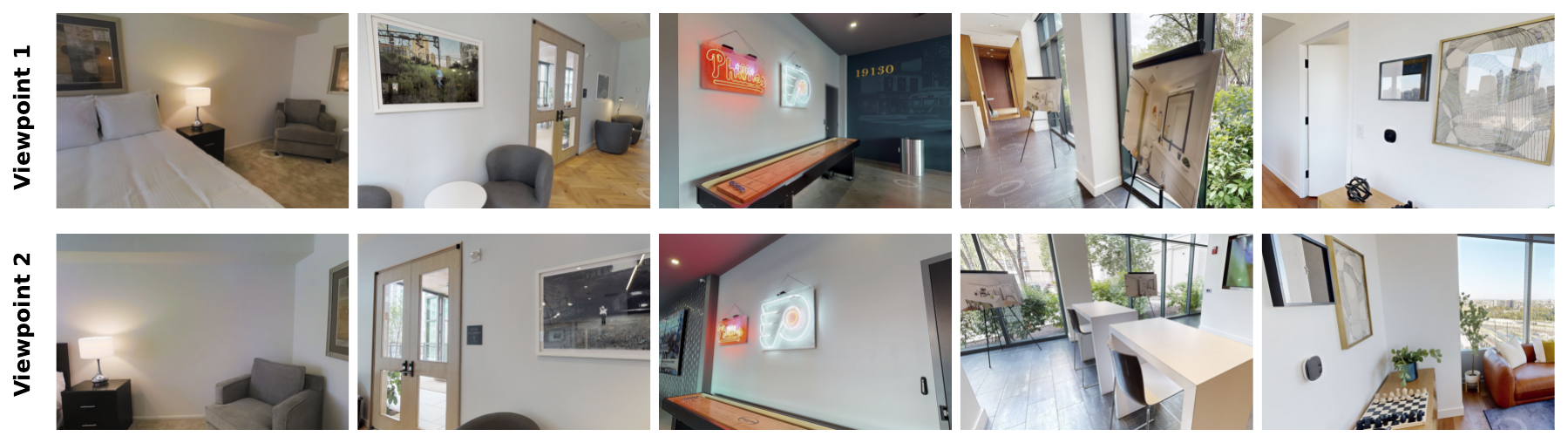}
   \caption{
Examples from the real-image viewpoint-consistency probe. Each column shows the same object pair from two viewpoints, with each view closer to a different object so that the nearer one projects larger. Images are taken from Matterport walkthroughs of real staged apartments.
}
    \label{fig:real_image}
\end{figure}

\subsection{Extra Experiments}
\label{app:extra_experiments}
\subsubsection{Open-Ended Robustness Probe}
\label{app:open_ended}

To rule out multiple-choice shortcuts, we re-evaluate a stratified sample of 600 questions using an open-ended protocol. The model receives no answer options and responds with a single constrained word (e.g., \texttt{left}, \texttt{right}, \texttt{same}, or \texttt{unsure}). The sample includes 300 2D-3D conflict questions and 300 occlusion questions, evenly split between answerable partial-occlusion and unanswerable full-occlusion cases.
As shown in Table~\ref{tab:open_ended}, Gemini-2.5-Flash shows nearly the same behavior under both protocols. Occlusion performance is unchanged, and under perspective conflict, the model still predominantly follows the projected 2D relation while rarely abstaining. 3D-consistent responses are slightly higher in the open-ended setting, suggesting that the multiple-choice format does not inflate performance. We additionally shuffle answer options for every question, yielding a balanced gold-label distribution.

\begin{table}[t]
\centering
\small
\caption{
Protocol robustness for Gemini-2.5-Flash. MC denotes multiple-choice evaluation, OE denotes open-ended evaluation without answer options, and Text denotes text-only evaluation without the image. 
}
\label{tab:open_ended}
\begin{tabular}{@{}llccc@{}}
\toprule
Setting & Metric & MC & OE & Text \\
\midrule

\multirow{3}{*}{Perspective conflict}
& 3D-Cons.$\uparrow$   & 33.7 & 41.3 & 0.0 \\
& Abstain              & 6.7  & 2.4  & 79.3 \\
& 2D-Foll.$\downarrow$ & 53.4 & 51.0 & 12.1 \\
\midrule

\multirow{2}{*}{Occlusion}
& Ans.$\uparrow$       & 49.1 & 49.3 & 20.0 \\
& Unans.$\uparrow$     & 56.0 & 56.7 & 49.3 \\
\bottomrule
\end{tabular}
\end{table}

\begin{table}[t]
\centering
\small
\caption{
Abstention precision and recall.
For occlusion, the positive class is an unanswerable question (22.0\% base rate);
for perspective, it is a 2D--3D conflict view (46.6\% base rate).
}
\label{fig:risk_coverage}
\begin{tabular}{lcc|cc}
\toprule
& \multicolumn{2}{c|}{\textbf{Occlusion}}
& \multicolumn{2}{c}{\textbf{Perspective}} \\
\textbf{Model}
& \textbf{Abs-P} & \textbf{Abs-R}
& \textbf{Abs-P} & \textbf{Abs-R} \\
\midrule
GPT-4o           & 41.9 & 63.6 & 46.8 & 38.3 \\
GPT-5-mini       & 62.9 &  8.4 & 61.5 &  8.9 \\
GPT-5.4          & 52.5 & 21.1 & 48.2 & 16.7 \\
Gemini-2.5-Flash & 50.7 & 49.4 & 68.2 &  4.8 \\
Gemini-3-Flash   & 61.8 & 51.4 & 73.6 &  6.2 \\
Qwen2.5-VL-7B    & 45.3 & 42.0 & 33.1 & 27.1 \\
Qwen2.5-VL-32B   & 50.4 & 45.5 & 47.2 & 16.4 \\
InternVL3-8B     & 54.0 &  8.3 & 83.3 &  0.8 \\
\bottomrule
\end{tabular}
\end{table}

\subsubsection{Selective-Prediction Analysis}
\label{tab:abstention_pr}

We further analyze whether model abstention is aligned with the cases in which
visual evidence is insufficient or potentially misleading. \Cref{tab:abstention_pr}
reports abstention precision and recall, treating unanswerable occlusion cases
and 2D--3D conflict views as the positive classes, respectively. The results
reveal markedly different abstention behaviors across models. Some models
abstain relatively often and recover a larger fraction of the designated cases,
but with limited precision, while others abstain more selectively and achieve
higher precision at the cost of very low recall. No model achieves both
precision and recall above 55\% in either setting. These results suggest that
current models do not consistently align their abstention behavior with
challenging visual conditions. For occlusion, this reflects difficulty in
identifying genuinely unanswerable observations; for perspective, it indicates
that abstention is not systematically triggered by views exhibiting a 2D--3D
conflict, rather than implying that all such views should necessarily be
abstained on.

\subsubsection{Details of the Representation and Attention Analyses}
\label{app:probe_attn}

\noindent\textbf{Data.}
Both analyses use the wall-unequal subset with Qwen2.5-VL-7B-Instruct.
For 270 conflict views and 81 clean reference views, we analytically project
the 3D mesh corners through the render camera to obtain image-plane object
boxes, together with mesh-derived physical-size labels. Among the 270 conflict
views, 198 are \emph{strict reversals}, where the physically smaller object
projects larger along the queried dimension; these are used for the attention
analysis.

\noindent\textbf{Linear probe.}
We freeze the model and extract features from vision-tower layers
(vis8--vis32) and LM layers (lm0--lm28). Object features are mean-pooled
within the left and right bounding boxes, and the probe input is
$[\mathbf{f}_L;\mathbf{f}_R;\mathbf{f}_L-\mathbf{f}_R]$.
A standardized logistic regression ($C{=}0.05$) predicts either the physical
3D relation or projected 2D relation using scene-level 5-fold cross-validation
(majority baseline $\approx$53\%). Because the two labels are complementary
on strict reversals, probes are trained on the decorrelated conflict set.
LoRA does not modify the vision tower, so vision-layer probe results remain
unchanged after fine-tuning. Probe accuracy measures linear decodability and
does not by itself imply that the model uses the encoded relation.

\noindent\textbf{Answer-token attention.}
Using the standard multiple-choice prompt, we record the first answer token's
attention to image tokens at each LM layer, averaged over heads with
attention-sink tokens removed. For each object, attention density is defined
as attention mass inside its bounding box divided by the box-area fraction,
controlling for projected object size. We report the density ratio between the
projected-larger and truly-larger objects for both the base and LoRA models.
For projection-following answers ($n{=}117$), the bias reaches
1.84$\times$ in middle layers and remains 1.54$\times$ in later layers,
whereas for 3D-consistent answers ($n{=}13$), late-layer attention returns to
near parity (0.99$\times$). This suggests, correlationally, that correcting the
misleading projection may require a late-stage shift in attention.

\begin{table}[t]
\centering
\caption{
Effect of LoRA fine-tuning under different observational uncertainty conditions. we report fine-tuning results on a held-out test split separate from the main benchmark evaluation. 
}
\label{tab:lora}
\small
\setlength{\tabcolsep}{4.5pt}
\begin{tabular}{lcc|cccc}
\toprule
& \multicolumn{2}{c|}{\textbf{Occlusion}}
& \multicolumn{4}{c}{\textbf{Perspective Ambiguity}} \\
\textbf{Model}
& \textbf{Ans.}$\uparrow$
& \textbf{Unans.}$\uparrow$
& \textbf{Ans.}$\uparrow$
& \textbf{3D-Cons.}$\uparrow$
& \textbf{Abstain}
& \textbf{2D-Foll.}$\downarrow$ \\
\midrule
Base (Qwen2.5-VL-7B)
& 53.1 & 69.3
& 15.9 & 3.4 & 37.9 & 58.6 \\

LoRA-Fintuned & \textbf{74.6} & \textbf{94.7} & \textbf{79.4} & \textbf{13.8} & 86.2 & \textbf{0.0} \\
\bottomrule
\end{tabular}
\end{table}

\begin{table}[t]
\centering
\small
\setlength{\tabcolsep}{7pt}
\renewcommand{\arraystretch}{1.05}
\caption{
\textbf{Effect of structured prompting on occlusion.}
Structured prompting improves unanswerable accuracy, but often reduces accuracy on answerable cases.
}
\label{tab:prompting}
\begin{tabular}{llcc}
\toprule
\textbf{Model} & \textbf{Prompt} & \textbf{Occ-Ans.} & \textbf{Occ-Unans.} \\
\midrule
\multirow{2}{*}{GPT-5-mini}
& Standard   & 66.2 & 8.4 \\
& Structured & 54.7 & 30.4 \\
\midrule
\multirow{2}{*}{Gemini-2.5-Flash}
& Standard   & 56.8 & 49.4 \\
& Structured & 50.4 & 52.7 \\
\bottomrule
\end{tabular}
\end{table}

\subsection{Implementation Details of the Evidence Interventions}
\label{app:interventions}

We evaluate all interventions on the same 209 wall-unequal reversal views
using Gemini-2.5-Flash with temperature 0. The model answers each comparison
with exactly one word from
\{\texttt{left}, \texttt{right}, \texttt{same}, \texttt{unsure}\}.
We compare four input conditions:
\textbf{Base:}The distorted RGB view only. \textbf{Depth as text:}
The same RGB view, augmented with the ground-truth camera distances of the
left and right objects.
\textbf{Depth map:}
The RGB view together with a pixel-aligned depth map estimated using
Depth-Anything-V2-Small. The depth map uses the \texttt{turbo} colormap with
per-image normalization, and the prompt specifies that red indicates closer
regions and blue indicates farther regions.
\textbf{Reference view:}
The distorted RGB view together with an additional reference view of the same
object pair. The question remains about the physical size relation in the
distorted view.
All conditions use the same question wording and differ only in the additional
evidence provided.

\subsubsection{Real-Image Viewpoint-Consistency Probe}
\label{app:real}

\noindent\textbf{Image sources.}
We collect real indoor images from two sources: (i) Matterport-based virtual
walkthroughs of real rental units from public apartment listings, and
(ii) self-captured photographs of household objects. Matterport tours provide
dense 3D scans of real interiors and allow controlled viewpoint changes similar
to our simulated setting. Platform watermarks and UI overlays are cropped from
all screenshots.

\noindent\textbf{Collection protocol.}
Each example contains two same-category objects photographed from two
viewpoints, with each view closer to a different object so that the nearer
object projects substantially larger. The model answers the same physical-size
comparison question independently from each view, with an abstain option.
Because the physical size relation is invariant to viewpoint, committed answers
that imply different physical relations across the two views constitute a
logical contradiction and certify at least one error without requiring
ground-truth size measurements. Abstentions are not counted as contradictions,
making the contradiction rate a conservative lower bound on error. When the
two viewpoints reverse image left/right, predictions are mapped to physical
object identity before comparison.

\noindent\textbf{Results.}
On 35 valid pairs without an obvious matched-set prior,
Gemini-2.5-Flash produces contradictory size judgments across viewpoints on
23 pairs (66\%). In every contradictory case, the model favors the object that
projects larger in the corresponding view. Because physical size is invariant
to viewpoint, these contradictions certify that at least one prediction is
incorrect, providing a conservative, measurement-free estimate of real-image
failure.

\subsection{Licenses and External Assets}
\label{app:license}
We use AI2-THOR (Apache 2.0) for simulation, open-source models (e.g., Qwen2.5-VL, InternVL3) under their respective licenses, and proprietary models (e.g., GPT and Gemini) via official APIs in accordance with their terms of service. 

\subsection{Broader Impact}
\label{app:borader impact}
This work studies observational uncertainty in vision-language models and highlights their tendency to produce confident spatial reasoning under incomplete or misleading observations. Improving awareness of unreliable visual evidence may benefit reliability-critical applications such as embodied agents and robotic systems. At the same time, our findings suggest that current models can make overconfident decisions under ambiguous viewpoints, which may lead to unsafe behaviors if deployed without appropriate safeguards. We hope this work encourages future research on uncertainty-aware and more reliable multimodal reasoning systems.

\subsection{Limitation}
\label{app:limitation}
Our framework relies on controlled synthetic 3D environments, which enable systematic control of observational conditions but may not fully capture the complexity and diversity of real-world scenes. In addition, our work focuses on observational uncertainty arising from occlusion and ambiguous viewpoints. While these settings isolate important challenges in spatial reasoning, real embodied environments may involve more complex and dynamic sources of uncertainty, such as motion, temporal changes, or sensor noise.
Besides, our evaluation focuses on single-step viewpoint assessment, rather than interactive exploration. Extending observational awareness to long-horizon embodied decision-making remains an important direction for future work.

\end{document}